\documentclass[letterpaper]{article} %
\usepackage{aaai2026}  %
\usepackage{times}  %
\usepackage{helvet}  %
\usepackage{courier}  %
\usepackage[hyphens]{url}  %
\usepackage{graphicx} %
\usepackage{natbib}  %
\usepackage{caption} %
\usepackage{algorithm}
\usepackage{algcompatible}

\usepackage{tabularx} 
\usepackage{booktabs}
\usepackage{array} 

\usepackage{newfloat}
\usepackage{listings}
\DeclareCaptionStyle{ruled}{labelfont=normalfont,labelsep=colon,strut=off} %
\floatstyle{ruled}
\newfloat{listing}{tb}{lst}{}
\floatname{listing}{Listing}
\usepackage{tikz}
\usetikzlibrary{bayesnet}
\usetikzlibrary{arrows}
\usepackage{booktabs}
\usetikzlibrary{shapes,decorations,arrows,calc,arrows.meta,fit,positioning}
\tikzset{
    -Latex, auto, node distance = 0.5 cm and 0.5 cm, semithick,
    state/.style = {circle, draw, minimum width = 0.7 cm},
    const/.style = {minimum width = 0.7 cm},
    inter/.style = {rectangle, draw, minimum width = 0.7 cm, minimum height = 0.7 cm},
    point/.style = {circle, draw, inner sep = 0.04cm, fill, node contents = {}},
    bidirected/.style = {Latex-Latex,dashed},
    el/.style = {inner sep=2pt, align=left, sloped}
}

\definecolor{SoftGreen}{HTML}{C9E7C9}

\newcommand{\Acal}{\mathcal{A}}

\newcommand{\Dcal}{\mathcal{D}}

\newcommand{\Fcal}{\mathcal{F}}
\newcommand{\Gcal}{\mathcal{G}}

\newcommand{\Ical}{\mathcal{I}}

\newcommand{\Mcal}{\mathcal{M}}
\newcommand{\Ncal}{\mathcal{N}}

\newcommand{\Scal}{\mathcal{S}}

\newcommand{\Ibb}{ \mathbb{I} }

\newcommand{\doit}{ {   \operatorname{do} } }

\newcommand{\zf}{z^F}
\newcommand{\zcf}{z^{CF}}
\newcommand{\zscf}{z^{\mathrm{SCF}}}
\newcommand{\zpscf}{z^{\mathrm{SPCF}}}

\newcommand{\cmark}{\ding{51}}%
\newcommand{\xmark}{\ding{55}}%
\usepackage{subcaption}
\usepackage{pifont}
\usepackage{amsmath}
\usepackage{amsfonts} %
\usepackage{amsthm}
\usepackage{multirow}
\usepackage{booktabs} 
\usepackage{array}
\usepackage[makeroom]{cancel}
\usetikzlibrary{bayesnet}
\usetikzlibrary{arrows}
\usepackage{listings}
\usepackage{adjustbox}
\usepackage{xcolor, soul}
\usepackage{algpseudocode}
\usepackage{makecell}

\usepackage{subcaption}
\usepackage{paralist}
\usepackage{adjustbox}

\usepackage{fontawesome5} %
\usetikzlibrary{shapes.geometric, arrows, calc}

\tikzstyle{startstop} = [circle, minimum size=1.5cm, text centered, draw=black, fill=white]
\tikzstyle{process} = [circle, minimum size=1.5cm, text centered, draw=black, fill=white]
\tikzstyle{arrow} = [thick,->,>=stealth, draw=black]
\tikzstyle{dashedarrow} = [thick,dashed,->,>=stealth, draw=black]

\title{A Causal Framework to Measure and Mitigate Non-binary Treatment Discrimination}
\author{
    Ayan Majumdar\equalcontrib\textsuperscript{\rm 1,2},
    Deborah D. Kanubala\equalcontrib\textsuperscript{\rm 2},
    Kavya Gupta\textsuperscript{\rm 2},
    Isabel Valera\textsuperscript{\rm 2}
}
\affiliations{
    \textsuperscript{\rm 1}Max Planck Institute for Software Systems, Saarbr\"ucken, Germany\\
    \textsuperscript{\rm 2}Saarland University, Saarbr\"ucken, Germany\\
    ayanm@mpi-sws.org, \{kanubala, gupta, ivalera\}@cs.uni-saarland.de
}

\begin{document}

\maketitle

\begin{abstract}
Fairness studies of algorithmic decision-making systems often simplify complex decision processes, such as bail or lending decisions, into binary classification tasks (e.g., approve or not approve). However, these approaches overlook that such decisions are not inherently binary; they also involve non-binary treatment decisions (e.g., loan or bail terms) that can influence the downstream outcomes (e.g., loan repayment or reoffending). We argue that treatment decisions are integral to the decision-making process and, therefore, should be central to fairness analyses.
Consequently, we propose a causal framework that extends and complements existing fairness notions by explicitly distinguishing between decision-subjects’ covariates and the treatment decisions. 
Our framework leverages path-specific counterfactual reasoning to: 
(i) measure treatment disparity and its downstream effects in historical data; and (ii) mitigate the impact of past unfair treatment decisions when automating decision-making. We use our framework to empirically analyze four widely used loan approval datasets to reveal potential disparity in non-binary treatment decisions and their discriminatory impact on outcomes, highlighting the need to incorporate treatment decisions in fairness assessments. Finally, by intervening in treatment decisions, we show that our framework effectively mitigates treatment discrimination from historical loan approval data to ensure fair risk score estimation and (non-binary) decision-making processes that benefit all stakeholders. 
\end{abstract}

\begin{links}
    \link{Code}{https://github.com/ayanmaj92/fair-nonbin-treat}
    \link{Extended version}{https://arxiv.org/abs/2503.22454}
\end{links}

\section{Introduction}\label{sec:intro}
\begin{figure}[t]
    \centering
    \includegraphics[width=1.0\linewidth]{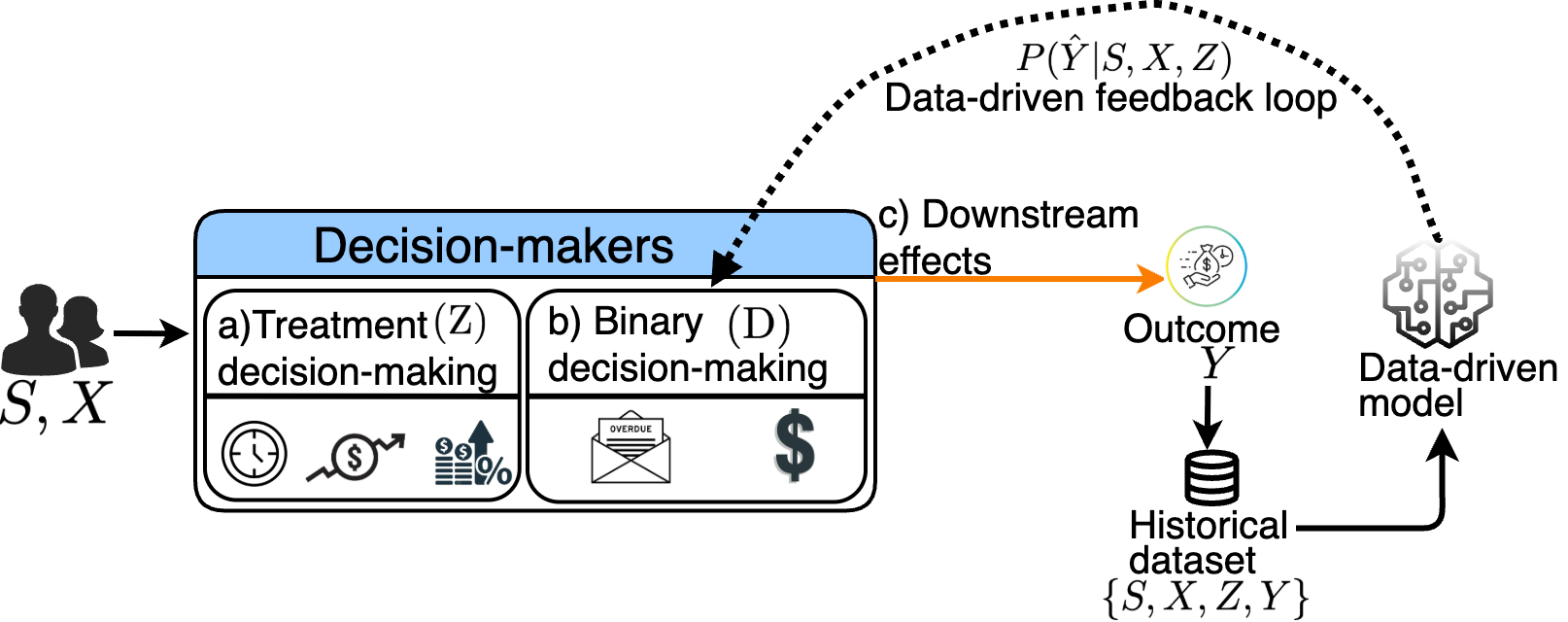}
   \caption{Illustration of data-driven decision-making pipelines. Parts (a)–(c) highlight the three common simplifications in existing fairness analyses.}
    \label{fig:pipeline}
\end{figure}

{Data-driven systems are increasingly used to automate decisions in domains such as finance, healthcare, and criminal justice~\cite{almheiri2023automated, moscato2021benchmark, habehh2021machine, dieterich2016compas}. These systems typically reduce complex decisions to binary classification tasks, for example, simply predicting for loan repayment or recidivism~\cite{mhasawade2024causal}}. However, \emph{real-world decisions are rarely as simple as a binary choice, they involve multiple, non-binary treatment decisions} which in turn affect outcomes. Fig.~\ref{fig:pipeline} illustrates the data-driven decision-making pipeline, which involves both (a) treatment decisions, (b) binary decisions,  and (c) their downstream effects.

Despite growing research on fairness in algorithmic decision-making, existing analyses simplify complex decision-making mechanisms in three key ways. First, many studies~\cite{zafar2017fairness, hardt2016equality, corbett2023measure} include non-binary treatments as covariates of the decision-subject (see covariate and treatment examples in Appendix Table~\ref{table:identification_of_features}), restricting fairness analyses to binary decisions (Fig.~\ref{fig:pipeline}b). This obscures the fact that such decisions are in control of the decision-makers. As a result, potential sources of discrimination may be masked, limiting the scope of fairness assessments. Second, some works~\cite{madras2019fairness, coston2020counterfactual, plecko2023causal} account for treatment decisions (Fig.~\ref{fig:pipeline}a) but assume they are binary, ignoring their often non-binary nature (e.g., loan or bail amounts). Finally, while a few studies do focus on disparities in non-binary treatment decisions, e.g., in lending~\cite{agier2013microfinance, escalante2018looking, alesina2013women}, they only provide empirical insights into disparities in treatment but overlook the downstream effects these treatment disparities may have on outcomes (Fig.~\ref{fig:pipeline}c).

In this paper, we bridge the gaps in the fairness literature by extending fairness analyses to multiple, non-binary treatment decisions and their downstream outcome effects. We propose a causal framework that explicitly distinguishes between a decision-subject's covariates and the treatment decisions made by a decision-maker. Our framework enables us to: (i) measure disparities in non-binary treatment decisions induced by the sensitive attributes, and (ii) quantify how these disparities propagate to outcomes, thereby inducing \textit{treatment discrimination}. Furthermore, we discuss how the proposed measures of treatment disparities and their downstream effects relate and complement existing fairness notions in binary decision-making.

{To make our framework actionable, we introduce a practical approach for generating fairness-aware counterfactual data, leveraging recent advances in causal inference~\cite{javaloy2024causal}. Unlike prior methods, which often relied on restrictive assumptions or computationally intensive procedures~\cite{nabi2018fair, chiappa2019path}, our approach is broadly applicable and scalable. Moreover, it supports \emph{treatment discrimination mitigation} via \emph{pre-processing} techniques. These techniques generate \emph{treatment-fair counterfactual datasets}, enabling fairer non-binary automated decisions and more equitable risk score estimation to inform future decision-making.}

We validate our framework using four real-world lending datasets across two decision-making scenarios and empirically show that: (i) historical data encode treatment disparities that fair binary predictions fail to correct for owing to downstream effects of treatment decisions; and (ii) fairer treatment decisions yield more optimal outcomes, benefiting all stakeholders.
Our findings thus highlight the need to address {treatment decision disparities} to achieve fairer outcomes, bridging a key gap in fairness research, and \emph{laying the foundation for more holistic fairness assessments and mitigation efforts.}
Thus, our analysis opens up avenues for future work, such as developing fair data-driven approaches for jointly making treatment and binary decisions.

\section{Background} \label{sec:background}

This section introduces the key causal concepts we leverage in our framework for studying treatment decision disparities.

\textbf{Structural causal models (SCMs).}
{
 An SCM~\cite{pearl2000models} encodes the \textit{cause-effect relations} among observed features $V$ as $\Mcal = (V, U, F)$, where $U$ is a set of mutually independent, unobserved exogenous variables, and $F$ is a set of functions such that each feature $V_i \in V$ is given by $V_i = F_i(\operatorname{pa}(V_i), U_i)$, with $\operatorname{pa}(V_i) \subseteq V$ denoting the causal parents of $V_i$. The structure of these equations induces a directed acyclic graph (DAG) $\Gcal$ over $V$, as seen in Fig.~\ref{fig:reformulated-graph}.
}

\textbf{Performing interventions.}
{
Interventions are external actions that forcibly fix a feature’s value, breaking its usual dependence on causal parents~\cite{pearl2000models}. In SCMs, such \emph{hypothetical interventions} are denoted by the $\doit$-operator. For instance, the intervention $\Ical = \doit(V_i = \alpha)$ replaces $V_i$’s structural equation with the constant assignment $V_i = \alpha$, producing a modified model $\Mcal^{\Ical}$. This \emph{intervened} SCM captures the changed causal structure and enables reasoning about the intervention’s downstream effects.
}

\begin{figure}[t]
    \centering
        \scalebox{0.85}{\begin{tikzpicture}[every node/.style={inner sep=0,outer sep=0}]
        \node[state, fill=white!60] (x) at (0,0) {$X$};
        \node[state, fill=white!60] (z) [right = 1.2 cm of x ] {$Z$};
    \node[state, fill=white!60] (y) [right = 1.2 cm of z ] {$Y$};
        \node[state, fill=white!60] (s) [left = 1.2 cm of x] {$S$};
    \draw[line width=1.15pt, dash pattern=on 2pt off 2pt, black] (s) -- (x);
    \draw[line width=1.15pt, dash pattern=on 5pt off 5pt, dash phase=4pt, red] (s) -- (x);
    
]    \draw[line width=1.15pt, dash pattern=on 2pt off 2pt, black] (x) -- (z);
    \draw[line width=1.15pt, dash pattern=on 5pt off 5pt, dash phase=4pt, red] (x) -- (z);
    
    \draw[line width=1.15pt, dash pattern=on 2pt off 2pt, black] (z) -- (y);
    \draw[line width=1.15pt, dash pattern=on 5pt off 5pt, dash phase=4pt, orange] (z) -- (y);
    \path (s) edge [thick, color=red, bend right] (z);
    \path (x) edge [thick, color=black, bend right] (y);
    \path (s) edge [thick, color=black, bend left] (y);
\end{tikzpicture}}
   \caption{
   Causal graph of the data generation process, distinguishing decision-subject covariates $X$ and treatment decisions $Z$, assuming positive binary decisions. Red paths (solid: direct, dotted: indirect) show potential disparate influence of sensitive $S$ on $Z$. The orange dashed path from $Z$ to $Y$ reflects possible discriminatory effects on outcomes.
   } 
  \label{fig:reformulated-graph}
\end{figure}

\textbf{Counterfactuals.} 
{
In the absence of hidden confounders, the intervened SCM $\Mcal^{\Ical}$ enables \emph{counterfactual analysis} for a specific individual with observed features $\vartheta^F$. This analysis answers the question: “What would the counterfactual features $\vartheta^{CF}$ be if we intervened with $\doit(V_i \rightarrow \alpha)$, all else remaining the same?” Computing counterfactuals involves three steps~\cite{pearl2000models}: (i) \emph{abduction} that infers the exogenous $u^F$ from $\vartheta^F$ using the original $\Mcal$; (ii) \emph{action} that intervenes using $\doit(V_i \rightarrow \alpha)$ to produce the intervened SCM $\Mcal^{\Ical}$; and (iii) \emph{prediction} that computes the counterfactual features $\vartheta^{CF}$ from $\Mcal^{\Ical}$ using $u^F$ and $V_i = \alpha$. Counterfactuals capture the \emph{total effect} of an intervention on \textit{downstream} causal features for the \textit{individual}. This framework is widely used in fairness research~\cite{kusner2017counterfactual} to quantify the total effect of sensitive attributes (e.g., gender) on binary decisions (e.g., granting loans).
}

\textbf{Path-specific counterfactuals.} 
{
In certain scenarios, instead of focusing on total effects, we may want to understand how one feature influences another through \textit{specific causal pathways}. This can be done using \emph{path-specific counterfactuals}, which apply multiple $\doit$ operations to isolate effects along selected paths. Consider a causal structure involving variables $\langle V_i, V_j, V_k \rangle$ with two paths: a direct path $V_i \rightarrow V_k$ and an indirect path $V_i \rightarrow V_j \rightarrow V_k$. For a factual instance $\vartheta^F = \langle v_i^F, v_j^F, v_k^F \rangle$, we can compute the counterfactual effect of $\doit(V_i \rightarrow \alpha)$ along each path separately. To isolate the direct path $V_i \rightarrow V_k$, we block the indirect path by fixing $V_j$ to its factual value, yielding $v_k^{CF}(\doit(V_i \rightarrow \alpha), \doit(V_j \rightarrow v_j^F))$. To isolate the indirect path $V_i \rightarrow V_j \rightarrow V_k$, we fix $V_i = v_i^F$ for the direct path, but update $V_j$ based on the intervention using $v_k^{CF}(v_i^F, \doit(V_j \rightarrow v_j^{CF}(\doit(V_i \rightarrow \alpha))))$. This approach has been effective in fairness studies~\cite{chiappa2019path} to determine whether sensitive features influence binary decisions \emph{through potentially problematic causal paths}.
}

\theoremstyle{definition}
\newtheorem{definition}{Definition}

\section{Measuring Treatment Disparities}
\label{sec:measures}
In this section, we introduce our causal framework to measure non-binary treatment disparities and their downstream effects.
Our framework considers historical data $\Dcal$ generated by the causal graph $\Gcal$ (Fig.~\ref{fig:reformulated-graph}), where:
\begin{enumerate}
\item \emph{Treatment decisions $Z$ are assigned by decision-makers} (e.g., banks setting loan terms) and are causally influenced by decision-subject covariates $X$ (e.g., income, savings) and possibly the sensitive attribute $S$ (e.g., gender, race). This mechanism \textit{may exhibit} disparities via \emph{direct} (solid red) and \emph{indirect} (dashed red) effects of $S$.
\item \emph{Treatment decisions $Z$ causally affect outcomes $Y$} (e.g., loan terms affect repayments). Thus, $S$ may influence $Y$ indirectly through $Z$ (orange dashed line in Fig.~\ref{fig:reformulated-graph}), potentially propagating treatment disparities to outcomes.
\end{enumerate}
\textbf{Remarks.}
{
Although Fig.~\ref{fig:reformulated-graph} presents a single $S$ for simplicity, our framework supports modeling multiple as independent root nodes. We treat $X$ and $Z$ as \emph{multivariate blocks}, focusing on causal relations between blocks without making assumptions about causal structures within them. The directed edges in $\Gcal$ indicate the \emph{causal ordering}~\cite{javaloy2024causal} among feature groups, although some of these relations may have zero effect. To enable general analyses, we retain $S \rightarrow Y$ to account for direct effects in scenarios like healthcare.
}

\subsection{Disparities in Non-Binary Treatment Decisions}

{
Based on the causal graph in Fig.~\ref{fig:reformulated-graph}, we use the SCM framework to measure how $S$ influences treatment decisions $Z$, capturing both the total and the direct disparities. For simplicity and without loss of generality, we assume a binary sensitive attribute, $S \in \{0,1\}$.
\footnote{Discussions on possible extensions to multi-valued and multiple sensitive features in Appendix~\ref{app:def_multisens}.} 
Given a dataset $\mathcal{D} = \{s^F, x^F, z^F, y^F\}$ consisting of instances that received a positive decision under a historical binary predictive policy, we define disparity measures using a disparity function $\Delta$, which depends on the type of treatment variable, e.g., using value differences for continuous treatments.
}

\begin{definition}
\label{def:ttd} 
    \emph{Total Treatment Disparity} (TTD)
    against decision-subjects with factual sensitive feature $s^F$, measures the expected disparity between the treatment $z^F$ they received and the treatment $\zscf$ they would have received if they belonged to a different sensitive group, i.e.,  
    \begin{align}
    \label{eq:ttd}
        \begin{split}
        \operatorname{TTD} (s^F) &= \mathbb{E}_{\Dcal}\left( \Delta(\zscf, \zf)\right) ,\, \mathrm{where} \\
        \zscf &= z^{CF}(\doit (S \rightarrow (1-s^F)))
        \end{split}
    \end{align}

{Following~\citet{kusner2017counterfactual}, this notion captures the \emph{total effect} of $S$ on the treatment decisions $Z$. The \emph{sensitive counterfactual} treatment $\zscf$ reflects, e.g., the loan terms a female applicant would receive if they were male with counterfactually adjusted covariates like income.}
\end{definition}

\begin{definition}
\label{def:dtd}
    \emph{Direct Treatment Disparity} (DTD), against decision-subjects with factual sensitive feature $s^F$, measures the expected disparity between the treatment they received and the sensitive \emph{direct-path counterfactual treatment}, %
    i.e., 
    \begin{align}
    \label{eq:dtd}
    \begin{split}
        \operatorname{DTD}(s^F) &= \mathbb{E}_{\mathcal{D}} \left( \Delta(\zpscf, \zf) \right),\, \mathrm{where} \\
        \zpscf &= z^{CF}(\doit (S \rightarrow (1-s^F)), \doit (X \rightarrow x^F))
    \end{split}
    \end{align}

    {
    Following~\citet{nabi2018fair,chiappa2019path}, this captures the \emph{direct effect} of $S$ on $Z$. The \emph{sensitive direct-path counterfactual} $\zpscf$ reflects, e.g., the loan terms a female applicant would have received if they identified as male while maintaining their original covariates like income.
    }
\end{definition}

\textbf{Relation to fair binary predictions.}
{
Ensuring fairness in binary decisions determines who receives a positive decision (e.g., loan granted) but does not directly address disparities in more complex, non-binary treatment assignments. We illustrate this empirically in Sec.~\ref{sec:case-studies}. Still, drawing parallels with predictive fairness concepts can help interpret the implications of treatment disparities. For example, $\operatorname{DTD}$ captures disparities caused by the \textit{explicit use} of sensitive attributes $S$ in assigning $Z$. While this may be acceptable in settings like healthcare, it is generally inappropriate in domains like lending, echoing the notion of \emph{disparate treatment}~\cite{dwork2012fairness}. Similarly, zero $\operatorname{TTD}$ requires $Z$ to be independent of $S$, analogous to \emph{demographic parity}~\cite{dwork2012fairness}, which requires binary predictions to be independent of $S$. As with predictive fairness, the appropriateness of treatment parity depends on context. When the true outcome $Y$ is unobserved (e.g., hiring), treatment disparities may be considered discriminatory. In contrast, when $Y$ is observable and the ground-truth (e.g., loan repayment), separation-based fairness frameworks~\cite{hardt2016equality,zafar2017fairness} suggest that \textit{outcome differences may justify some disparities}. Hence, analyzing the downstream impact on $Y$ is essential for assessing the unfairness of treatment disparities.
}
\subsection{Downstream Effects of Treatment Disparities}

While Def.~\ref{def:ttd} and~\ref{def:dtd} help measure treatment disparities, it is crucial to assess the downstream impact of these disparities on the outcome $Y$ (e.g., when $Y$ is a ground truth as in lending) to determine if they are discriminatory. Next, we introduce two novel metrics to quantify the downstream outcome effects of total and direct treatment disparities.

\begin{definition}%
\label{def:tte_y}
\emph{Total Treatment Disparity Effect} {(TTD-E)} measures the downstream effect of the sensitive counterfactual treatment $\zscf$, computed in Eq.~\ref{eq:ttd}, as the probability of a \emph{label change} from the factual value $y^F$, i.e., 
\begin{align}
    \label{eq:tte_y}
    \begin{split}
   \operatorname{TTD-E} (s^F,y^F)
   &=  \mathbb{E}_{\mathcal{D}}\left[ \mathbb{I}\left(y^{CF}( \doit (Z \rightarrow \zscf) )  \neq y^F \right)\right],
    \end{split}
\end{align}
{where $\mathbb{I}$ is an indicator function. This measures how the outcome would change if the decision-maker applied the \emph{sensitive counterfactual} treatment $\zscf$, e.g., how a female applicant’s repayment would differ if her covariates stayed the same but treatment was adjusted by the \emph{total effect of} $S$.}

\end{definition}

\begin{definition}
\label{def:dte_y}
\emph{Direct Treatment Disparity Effect} {(DTD-E)}
measures the downstream effect of the \textit{sensitive path} counterfactual treatment $\zpscf$, computed in Eq.~\ref{eq:dtd}, as the probability of a  \emph{label change} from the factual value $y^F$, i.e., 
\begin{equation}
    \label{eq:dte_y}
    \operatorname{DTD-E} (s^F,y^F) 
     =  \mathbb{E}_{\mathcal{D}}\left[ \mathbb{I}\left(y^{CF}( \doit (Z \rightarrow \zpscf)) \neq y^F \right) \right]. 
\end{equation}
{This measures how the outcome would change if the decision-maker applied the \emph{sensitive direct-path counterfactual treatment} $\zpscf$, e.g., how a female applicant’s repayment would differ if her covariates stayed the same but only the treatment was adjusted by the direct effect of $S$.}
\end{definition}

\textbf{Relation to fair binary predictions.}
{
Ensuring parity in treatment decisions \textit{complements} parity in binary decisions, and fairness in predictions alone cannot eliminate unfair effects of treatment disparities on outcomes. However, connecting to predictive fairness concepts helps identify when treatment disparities lead to \textit{discrimination through unfair decision outcomes}. For example, under the notion of \emph{disparate treatment}~\cite{dwork2012fairness}, any direct use of sensitive attribute $S$ in assigning treatment $Z$ is problematic, and its impact on outcome $Y$ is considered discriminatory. However, in some contexts like healthcare, DTD effects may be justified if they benefit $Y$. Similarly, disparities between counterfactual and factual treatments ($Z^{\mathrm{SCF}}$ and $Z^F$) can cause unfairness in $Y$. In predictive fairness, separation~\cite{zafar2017fairness} allows decision disparities if they do not harm ground-truth outcomes. Likewise, treatment disparities (e.g., TTD) may be deemed discriminatory if they fail to improve $Y$, such as when factual treatment $\zf$ leads to worse outcomes than counterfactual treatment $\zscf$.
}

\subsection{Practical Implementation}
\label{sec:implementation}

{
Prior work~\cite{nabi2018fair,chiappa2019path} on path-specific counterfactuals for predictive fairness was either limited to linear settings or relied on imprecise approximations and regularization. To allow for a \textit{practical yet theoretically grounded} method for computing path-specific counterfactuals, we leverage \emph{causal normalizing flows} (CNF)~\cite{javaloy2024causal}, which have shown accurate causal estimates in societal decision-making settings~\cite{majumdar2024carma}.
}

\paragraph{Background on CNF.}
\label{sec:flows_bck}
{
Given factual data $\mathcal{D} =  \{ s^F, x^F, z^F, y^F \}$ and causal graph $\Gcal$, CNF approximates the unknown SCM $\mathcal{M}$ using a single invertible normalizing flow model $T_{\psi}$. CNFs support \emph{partial causal graphs}, where some features are grouped into \emph{multivariate blocks}, requiring only a \emph{causal ordering} across blocks. CNFs can also flexibly learn from data if certain causal relations have \emph{zero effect}. Once trained, CNF estimates counterfactuals for a given factual instance $\vartheta^F = \{v_1^F, \ldots, v_K^F\}$ under an intervention $\doit(V_j \rightarrow \alpha)$ through three steps: (i)~\emph{Abduction}; infer exogenous variables as $u^F = T_\psi(\vartheta^F)$; {(ii)~\emph{Action}}, modify $u^F$ to reflect the intervention, yielding $u^{CF} = \{u^F_{1:j-1}, T_\psi(\doit(V_j \rightarrow \alpha)), u^F_{j+1:K}\}$; and,  {(iii)~\emph{Prediction}}, estimate counterfactual $\vartheta^{CF} = T_\psi^{-1}(u^{CF})$.
See Appendix~\ref{app:CNF} for extended details. Next, we show how to adapt CNF to estimate treatment disparities.
}

\paragraph{Estimating counterfactual treatments and outcomes.}
We can use a causal normalizing flows model trained using factual $\Dcal$ to compute counterfactual quantities for some factual $\vartheta^F \in \Dcal$.
Importantly, we can readily utilize the \emph{existing} counterfactual estimation method shown above to compute the \emph{sensitive total treatment counterfactual} $\zscf$ as: 
\begin{align}
    \zscf = T^{-1}_\psi (\langle T_{\psi} ( \doit(s \rightarrow (1-s^F))), {u}^F_{X}, {u}^F_Z, {u}^F_Y \rangle)
\end{align}
We can also directly use the existing CNF approximation process to compute the \emph{counterfactual labels} in Def.~\ref{eq:tte_y}~and~\ref{eq:dte_y} resulting from a treatment intervention $\doit (Z \rightarrow \hat{z})$, as: 
\begin{align}
    y^{CF}(\doit (Z \rightarrow \hat{z})) = T^{-1}_\psi (\langle {u}^F_S, {u}^F_{X}, T_\psi ( \doit(z \rightarrow \hat{z})), {u}^F_Y \rangle),
\end{align}
 where $\hat{z}$ denotes respectively $\zscf$ from Def.~\ref{eq:ttd}, and  $\zpscf$  from Def.~\ref{eq:dtd}. 
 However, note that to compute the effect of the counterfactual treatments on the outcome, we need to perform counterfactual approximations \emph{in two steps}, where the first step measures the counterfactual treatment and the second step measures the effect on the outcome. 

\paragraph{Extension for path-specific disparity measures.}
To analyze our path-specific measures in Def.~\ref{eq:dtd}~and~\ref{eq:dte_y}, we cannot directly apply the existing CNF counterfactual estimation process since it does not work for multiple sequential interventions.
Moreover, since we want to compute path-specific measures in non-linear settings, we cannot directly apply the existing linear approach in~\cite{chiappa2019path} either.
Hence, we provide a \emph{novel procedure} that \emph{extends} the existing non-linear causal normalizing flows counterfactual approach to approximate \emph{path-specific counterfactuals}.

For computing $\zpscf$ in Def.~\ref{def:dtd} and~\ref{def:dte_y}, we need to perform \textit{sequential interventions} in the CNF \emph{following the causal ordering} of the features to obtain:
\begin{align}
\begin{split}
    &\zpscf = T^{-1}_{\psi}(\langle {u}^{CF}_S, \hat{{u}}_X^{F}, {u}^F_Z,{u}^F_Y \rangle),\, \mathrm{where} \\
    &{u}^{CF}_S = T_{\psi}(\doit(S \rightarrow (1-s^F))),\, \mathrm{and} \,\, \\
   &  \hat{{u}}_X^{F} = T_{\psi}((1-s^F), \doit(X \rightarrow x^F))
\end{split}
\end{align}
Hence, to compute $\zpscf$, we perform two interventions following the causal ordering. 
First we use $\doit(S\rightarrow (1-s^F))$ on $\vartheta^F$ to get $\vartheta^{\mathrm{SCF}}$.
Then, we use $\doit(X \rightarrow x^F)$ on $\vartheta^{\mathrm{SCF}}$, combine the different exogenous $u$ from the separate intervention steps and generate $\zpscf$.
Further details and pseudocode regarding the different counterfactual computations using CNF can be found in Appendix~\ref{apx:pseudocodes}.

{
\paragraph{Assumptions and considerations.}
Our causal framework is designed to incorporate advances in causal generative modeling flexibly, but its current CNF-based implementation relies on key assumptions. Our framework assumes \textit{overlap}, i.e., all relevant treatment-covariate combinations have non-zero support, a condition that generally holds in practice (Appendix~\ref{apx:assumptions}). Consistent with prior work~\cite{khemakhem2021causal,sanchez2022vaca,kusner2017counterfactual,chiappa2019path}, we also assume \textit{no hidden confounders} and \textit{full observability} of treatments, covariates, and sensitive attributes from the \textit{decision-maker’s perspective}. 
However, in real-world settings, decision-makers \emph{may not observe} all covariates impacting the outcome, introducing potential confounding. However, this \emph{would only impact} the direct disparity measures (Definitions~\ref{def:dtd} and~\ref{def:dte_y}) since they need interventions on $X$.
Importantly, since decision-makers only intervene on treatment decisions in our framework, robustness can be practically tested through small-scale interventional studies on subpopulations. Future work can also explore integrating recent advancements for tackling confounding~\cite{almodovar2025decaflow}.
We provide a more detailed discussion in Appendix~\ref{apx:assumptions}.
Appendix~\ref{app:synthetic} presents synthetic data experiments \textit{validating the framework's effectiveness against an oracle}. 

}
\section{Mitigating Historical Treatment Unfairness}
\label{sec:mitigation}
{Our framework identifies when historical data reflects discriminatory treatment policies, i.e., disparities in treatments $Z$ that are not justified by corresponding benefits in ground-truth outcomes $Y$. To prevent such biases from being perpetuated in data-driven decision-making, we propose an automated pre-processing procedure. Although pre-processing approaches may lack formal guarantees, they remain a practical and effective mitigation strategy~\cite{mutlu2022contrastive}. Our method augments the biased dataset $\Dcal$ to produce a fairer version, $\Dcal^{\mathrm{fair}}$, enabling more equitable non-binary decisions (Sec.~\ref{sec: fair-nonbinary}). We further show that using $\Dcal^{\mathrm{fair}}$ can mitigate the effects of treatment disparities on risk score estimation, a central component in many decision-making pipelines (Sec.~\ref{sec: fair-risk}).
}

\subsection{Fairness in Non-Binary Decision-Making}
\label{sec: fair-nonbinary}
{Automated decision-making systems often rely on historical data, including sensitive attributes, individual covariates, and past treatments to train classifiers that predict outcomes: \( h_y: \{S, X, Z\} \mapsto Y \). These predictions are used to make binary decisions, such as granting loans~\cite{corbett2017algorithmic}. However, biases in the decision-maker’s \emph{non-binary past treatment policy} (Def.~\ref{def:ttd}--\ref{def:dte_y}) can propagate into future automated decisions. As we show later in Sec.~\ref{sec:case-studies} and Appendix~\ref{app:additional_eval}, optimizing for predictive fairness~\cite{dwork2012fairness,hardt2016equality} \emph{does not eliminate treatment disparities or their downstream effects}.
}

{To prevent automated systems from reinforcing such disparities, we propose a pre-processing method that uses our causal framework to transform biased historical data $\mathcal{D}$ into a \emph{treatment-fair} dataset $\mathcal{D}^{\mathrm{fair}}$. We define a treatment policy $\pi(Z)$ that adjusts treatment assignments for disadvantaged individuals $S^F = s^-$, those who previously received worse treatment \emph{without benefiting in outcome} (Def.~\ref{def:tte_y}). For these cases, $\pi(Z)$ assigns the \emph{sensitive counterfactual} treatment $\zscf$ (Def.~\ref{def:ttd}). The fair dataset is then constructed as:
}
\begin{equation}
\begin{split}
\mathcal{D}^{\mathrm{fair}} &=  \left\{ s^F, x^F, z^F, y^F \right\}_{s = s^+} \bigcup \\
&\left\{ s^F, x^F, \zscf, y^{CF}(\doit (Z \rightarrow \zscf)) \right\}_{s = s^-}
\end{split}
\end{equation}
{For individuals in group $S^F = s^-$, we replace the historically assigned treatment with $\zscf$ and simulate the counterfactual outcome under this intervention (Def.~\ref{def:ttd} and~\ref{def:tte_y}). Despite its simplicity, $\pi(Z)$ corrects disparities without harming outcomes, ensuring $P\left[Y^{CF}(\doit (Z \rightarrow \zscf)) = 1 \mid S^F = s^- \right] \geq P\left[Y^F = 1 \mid S^F = s^- \right].$
The resulting $\mathcal{D}^{\mathrm{fair}}$ \emph{removes past disparate treatment and its effects}, enabling the training of predictive models that do not perpetuate such disparities.
}

\subsection{Fair Risk Score Estimation}
\label{sec: fair-risk}
{Risk scores are widely used in algorithmic decision-making to inform lending decisions~\cite{hurley2016credit, obermeyer2019dissecting}. For example, banks often rely on risk scores from third-party institutions such as FICO or SCHUFA to assess creditworthiness and justify loan terms~\cite{doroghazi2020fico, toh2023addressing, loqbox2025}.
These scores are typically estimated from historical data $\mathcal{D}$ and past decision outcomes~\cite{maiden2024fico, myfico2025}}
{as $R(X,Z) = P(Y = 0 \mid X, Z).$
However, disparities in prior non-binary treatment decisions can bias both outcomes and risk scores. Crucially, this bias may persist even if treatment decisions are excluded from the estimation, due to the causal effect of treatment $Z$ on outcome $Y$.}

{
Since the treatment decisions are beyond an applicant's control~\cite{kleinberg2016inherent,coston2020counterfactual}, we argue for mitigating the unfair effects of past treatment policies on risk score estimation. To this end, we formulate a fair risk score estimate as $R_\text{fair}(x,s) = \mathbb{E}_{z' \sim \pi(Z)}\left[P\left(y^{CF}(s^F, x^F, \doit (Z \rightarrow z')) = 0 \right)\right],$
where we use $\pi(Z)$ to marginalize out disparate effects of $Z$. To ensure the marginalization does not systematically disadvantage any group, we set $\pi(Z)=P_{\mathcal{D}^{\mathrm{fair}}}(Z \mid S=s^+)$, using treatments from the advantaged group. Treatment distribution overlap across groups (Appendix~\ref{apx:assumptions}) ensures our interventions remain realistic while removing disparate effects. %
}

\begin{table}[t!]
 \centering
\resizebox{0.95\columnwidth}{!}{
 \begin{tabular}{l c c c c c c }
    \toprule
     \multirow{3.5}{*}{\textbf{Data}} & \multirow{3.5}{*}{{\textbf{\begin{tabular}[c]{@{}l@{}}SCF\end{tabular}}}}  & \multirow{3.5}{*}{\textbf{Measure}} & \multicolumn{2}{c}{\textbf{Disparity}} & \multicolumn{2}{c}{\textbf{Effect($\%$)}}\\
    \cmidrule(lr){4-5} \cmidrule(lr){6-7}
    & & & \multicolumn{2}{c}{\textbf{Amount (K\$)}} &  \multicolumn{2}{c}{\boldmath{$Y^F= 0/1$ }}\\ %
    \cmidrule(lr){4-5} \cmidrule(lr){6-7} %
    & & & \textbf{Hist.} & \textbf{EOD} & \textbf{Hist.} & \textbf{EOD}\\
    \midrule
    \multirow{4}{*}{\textbf{NY}} & \multirow{2}{*}{F$\rightarrow$M} & TTD(-E) & $+30$ & $+28$ & $2.2/0.4$ & $4.8/0.1$ \\ %
    & & DTD(-E) & $+7$ & $+7$ & $2.5/0.2$ & $3.6/0.1$\\ %
    \cmidrule(lr){4-5} \cmidrule(lr){6-7} %
    & \multirow{2}{*}{M$\rightarrow$F} & TTD(-E) & $-30$ & $-28$ & $3.9/0.2$ & $2.6/0.0$ \\ %
    & & DTD(-E) & $-6$ & $-7$ &  $1.8/0.1$ & $2.6/0.0$\\ %
    \midrule
    \midrule
    \multirow{4}{*}{\textbf{TX}} & \multirow{2}{*}{F$\rightarrow$M} & TTD(-E) & $+19$ & $+21$ & $1.6/0.3$ & $0.0/0.0$ \\
    & & DTD(-E) & $+3$ & $+3$ & $1.6/0.1$ & $0.0/0.0$\\ %
  \cmidrule(lr){4-5} \cmidrule(lr){6-7} %
    & \multirow{2}{*}{M$\rightarrow$F} & TTD(-E) & $-20$ & $-21$ & $3.0/0.2$ & $0.0/0.0$\\
    & & DTD(-E) & $-4$ & $-3$ & $2.4/0.1$ & $0.0/0.0$ \\
    \bottomrule
 \end{tabular}
 }
  \caption{
 \textbf{Treatment disparities and outcome effects in S.1 (HMDA).} Disparity in median U.S.D. ($\$$) across sensitive counterfactuals (SCF), female (F) and male (M). $Y^F =0$: label changed from negative to positive agreement; $Y^F =1$: reverse. {\textbf{Hist.}: all test-set loans granted (past policy); \textbf{EOD}: use loan grants predicted by equalized odds predictor.}
 }
\label{tab:HMDA}
 \end{table}

\begin{table*}[t!]
\centering
\resizebox{0.88\textwidth}{!}{%
\begin{tabular}{l c c c c c c c c c c}
    \toprule
     \multirow{3.5}{*}{\textbf{Data}} & \multirow{3.5}{*}{{\textbf{\begin{tabular}[c]{@{}l@{}}SCF\end{tabular}}}}  & \multirow{3.5}{*}{\textbf{Measure}} & \multicolumn{4}{c}{\textbf{Disparity}} & \multicolumn{4}{c}{\textbf{Effect ($\%$)}}\\
   \cmidrule(lr){4-7}  \cmidrule(lr){8-11}
     & & & \multicolumn{2}{c}{\textbf{Annuity (INR)}} & \multicolumn{2}{c}{\textbf{Amount (INR)}} & 
     \multicolumn{2}{c}{\boldmath{$ Y^F= 0$}} & \multicolumn{2}{c}{\boldmath{$ Y^F= 1$}} \\
     \cmidrule(lr){4-5} \cmidrule(lr){6-7} \cmidrule(lr){8-9} \cmidrule(lr){10-11}
     & & & \textbf{Hist.} & \textbf{EOD} & \textbf{Hist.} & \textbf{EOD} & \textbf{Hist.} & \textbf{EOD} & \textbf{Hist.} & \textbf{EOD} \\
      \midrule
     \multirow{4}{*}{\textbf{Home Credit}} & \multirow{2}{*}{F $\rightarrow$ M} & TTD(-E) & $+1005.26$ & $+1031.86$ & $+7834.47$ & $+8606.90$ & $0.22$ & $0.18$ & $0.25$ & $0.13$ \\
    & & DTD(-E) & $+708.24$ & $+694.17$ & $-797.40$ & $-972.88$  & $0.22$ & $0.00$ & $0.08$ & $0.05$ \\
    \cmidrule(lr){4-5}  \cmidrule(lr){6-7} \cmidrule(lr){8-9} \cmidrule(lr){10-11}
    & \multirow{2}{*}{M $\rightarrow$ F} & TTD(-E) & $-957.37$ & $-924.41$ & $-6778.88$ & $-7310.10$ & $2.46$ & $3.08$ & $0.05$ & $0.04$\\
    & & DTD(-E) &  $-604.94$ & $-529.54$ & $+857.12$ & $+1069.00$ & $1.36$ & $1.54$ & $0.08$ & $0.06$ \\
     \midrule
     \midrule
     & & & \multicolumn{2}{c}{\textbf{Duration (months)}} & \multicolumn{2}{c}{\textbf{Amount (DM)}} & 
     \multicolumn{2}{c}{\boldmath{$ Y^F= 0$}} & \multicolumn{2}{c}{\boldmath{$ Y^F= 1$}} \\
     \cmidrule{4-11}
    \multirow{4}{*}{\textbf{German Credit}} & \multirow{2}{*}{F $\rightarrow$ M} & TTD(-E) & $+1.15$ & $0.94$ & $+272.15$ & $+189.89$ & $4.25$ & $0.00$ & $2.70$ & $3.64$ \\
    & & DTD(-E) & $+0.42$ & $+0.36$ & $+133.39$ & $+116.74$ & $4.25$ & $0.00$ & $1.35$ & $1.82$\\
    \cmidrule(lr){4-5}  \cmidrule(lr){6-7} \cmidrule(lr){8-9} \cmidrule(lr){10-11}
    & \multirow{2}{*}{M $\rightarrow$ F} & TTD(-E) & $-0.87$ & $-0.67$ & $-237.97$ & $-212.44$ & $12.33$ & $9.10$ & $0.97$ & $0.81$ \\
    & & DTD(-E) & $-0.58$ & $-0.42$ & $-189.68$ & $-189.68$ & $12.33$ & $4.54$ &  $0.48$ & $0.81$\\
    \bottomrule
\end{tabular}
}
\caption{
\textbf{Treatment disparities and outcome effects in S2}. For sensitive counterfactuals (SCF), female (F) and male (M), $Y^F = 0$: outcome changed defaulted $\rightarrow$ repaid, $Y^F = 1$ the reverse. Reporting median INR (Indian Rupees) and DM (Deutsche Marks). \textbf{Hist.} all test-loans were granted (past policy), \textbf{EOD} only uses test-loans granted by the equalized odds predictor.
}

\label{tab:loanrepayment-analysis}
\end{table*}
\section{Use Case: Lending Decisions} \label{sec:case-studies}
{In this section, we evaluate our framework by (i) auditing historical data to identify treatment disparities and their \emph{potentially discriminatory} effects on outcomes, and (ii) assessing the effectiveness of our mitigation strategies. We evaluate using four real-world lending datasets that span diverse geographic regions, loan structures, and two representative decision-making scenarios (experimental details in Appendix~\ref{app:setup} and additional results in Appendix~\ref{app:additional_eval}):}
\begin{itemize}
    \item \textbf{ Decision agreement as label (S.1):}
The \textit{ground-truth} $Y$ captures a decision's \emph{immediate outcome}, where $Y=1$ shows \emph{agreement on treatment terms}, e.g., a loan approved by the bank and terms accepted by the applicant. We evaluate this scenario using the 2017 HMDA dataset from New York and Texas~\cite{ffiec2022housingdata}.
\item \textbf{Decision outcome as label (S.2)}: The \textit{ground-truth} $Y$ captures a decision's \emph{downstream outcome}, e.g., $Y=1$ for loan repaid and $Y=0$ for default. We analyze this scenario using Home Credit~\cite{homecredit} and German Credit~\cite{hofmann1994statlog} data.

\end{itemize}
\subsection{Measuring Treatment Disparities}
We begin by auditing disparities in historical treatment assignments in test data. First, we examine disparities under historical binary decisions using the full test set (Hist.), where all individuals received positive decisions. We also evaluate the impact of predictive fairness by analyzing treatment disparities for positively predicted individuals under an equalized odds (EOD) predictor. 
Results for other predictors in Appendix~\ref{apx:mult-preds}, analysis of unsplit data in Appendix~\ref{apx:whole-data-analysis}.

\subsubsection{S.1: Agreement of the decision-making process as label.}

In both HMDA datasets, we follow the identification and pre-processing procedures for sensitive attributes, covariates, and treatment decisions outlined in~\cite{cooper2024arbitrariness}. 
We consider loan amount and preapproval status as treatment decisions within the bank's authority. Table~\ref{tab:HMDA} reports the treatment disparity analysis results.

\textbf{Takeaway. }
Although loan agreement rates do not differ significantly by gender, our analysis (ref. Hist) shows that banks assign more conservative loan terms to female applicants, suggesting \textit{potential treatment unfairness}. Offering higher loan amounts to women would \textit{not} reduce agreement rates, challenging the justification for such disparities.
We also observe that \textit{predictive fairness} (e.g., EOD) \textit{does not eliminate treatment disparities}. For example, Table~\ref{tab:HMDA} shows that in the NY dataset, an EOD-compliant predictor still yields disparities: \emph{females, if treated as males, would have received \$27K higher loan amounts} without any negative impact on $Y$, indicating \textit{potential treatment unfairness}. These differences are largely explained by applicant covariates, with minimal direct disparity observed. 

Our observations raise important questions about the rationale behind these disparate treatment decisions. Specifically, we ask: \textit{Are disparities in loan terms for female applicants justified by downstream outcomes, such as repayment behavior, and thus defensible as a business necessity?} 
This motivates our second scenario examining treatment disparities, repayment outcomes, and potential discrimination.

\subsubsection{S.2: Decision outcome as label.}
In the Home Credit dataset, we analyze two treatment decisions: loan amount and annuity amount, while in the German Credit dataset, we examine three treatment decisions: loan amount, loan duration, and installment rate 
with results summarized in Table~\ref{tab:loanrepayment-analysis}.

\textbf{Takeaway.}
In Scenario S.2, our results show female applicants experience negative treatment disparities \emph{not explained} by downstream outcomes. Conversely, while males receive \emph{perceived positive} treatment, they are ultimately \textit{negatively impacted} by its downstream effects.
Notably, our findings align with similar insights observed for German Credit in a simpler linear setting~\cite{kanubala2024fairness}, reinforcing the validity of our results. Despite appearing favorable, loan terms offered to certain males may increase their default risks and lower their creditworthiness over time. 
Applying the treatment decisions of their female counterparts to male applicants could mitigate this discrimination and improve repayment outcomes. 
Importantly, this scenario also further illustrates that \emph{predictive fairness alone does not mitigate treatment disparities or their effects on outcomes}. In German Credit, 9.1\% of male applicants who received loans from the EOD predictor but later defaulted \textit{would have repaid under counterfactual treatments}.

\subsection{Mitigating Treatment Discrimination} 
\label{sec:exp_mitigation-data}
Our results in Table~\ref{tab:loanrepayment-analysis} show that banks exhibit negative treatment disparity against female applicants. 
However, males are \textit{potentially discriminated against} since they experience a negative downstream effect on repayment ability, and treating them as females improves repayment performance. Hence, \textit{treating all applicants conservatively as females} may enhance the overall repayment ability of borrowers, benefiting both the banks and the applicants. 

\begin{table}[t!]
    \centering 
    \small
    \resizebox{0.99\columnwidth}{!}{%
    \begin{tabular}{llcc}
    \toprule 

    \textbf{Dataset} & \multicolumn{1}{l}{\textbf{
    Group}} & \multicolumn{1}{l}{\textbf{LGD (INR/DM)}}   & \multicolumn{1}{l}{\textbf{ESI (INR/DM)}} \\
    \midrule 
    \multirow{3}{*}{\textbf{\begin{tabular}[c]{@{}c@{}}Home \\ Credit\end{tabular}}} & Female &  $586692.67$  & $ 290208.73$ \\
    &  Male ($\Dcal$) &  $576590.35$  & $422114.31$ \\
    & Male ($\Dcal^\mathrm{fair}$) &  $\textbf{568157.43}$  & $\textbf{391110.73}$ \\
    \midrule
    \multirow{3}{*}{\textbf{\begin{tabular}[c]{@{}c@{}}German \\Credit\end{tabular}}}& Female &  $1220.51$  & $ 312.13$ \\
    & Male ($\Dcal$) &  $1163.88$  & $495.65$\\
    & Male ($\Dcal^\mathrm{fair}$) &  $\textbf{940.03}$ & $\textbf{393.86}$ \\
    \bottomrule
    \end{tabular}
    }
     \caption{ 
\textbf{Fairness analysis in S.2}, comparing bank's LGD and applicant's ESI (both: lower better) for various policies.
$\Dcal$ is factual data,  in $\Dcal^\mathrm{fair}$ all applicants treated as females.
}
    \label{tab:losses}
\end{table}
\textbf{Fair non-binary decision-making.} Following Sec.~\ref{sec: fair-nonbinary}, we examine the creation of a \emph{treatment-fair} dataset aimed at automating future fair non-binary decision-making tasks. Based on our Scenario S.2 analyses, we identify males as the group needing adjusted treatment to generate our fair dataset, $\Dcal^\mathrm{fair}$. 
To understand the impact of $\Dcal^\mathrm{fair}$, we aim to compare the fair dataset with the original (unfair) dataset on two key metrics: the bank's Loss Given Default ($\operatorname{LGD}$)~\cite{schuermann2004we} and the applicant's Expected Simple Interest ($\operatorname{ESI}$)~\cite{ross2014fundamentals}. 
Appendix~\ref{sec:append-losses} contains further details on stakeholder losses.

\textbf{Takeaway.} 
Table~\ref{tab:losses} shows at the \textit{data-level} how adjusting male treatments reduces bias and benefits stakeholders by lowering \(\operatorname{LGD}\) for banks \textit{and} \(\operatorname{ESI}\) for male borrowers, highlighting the value of fairer treatment. Appendix~\ref{apx:train-fair} further confirms that training binary predictors on \(\mathcal{D}^{\mathrm{fair}}\) \emph{improves} outcome utilities while preserving predictive performance.

\begin{figure}[t!]
    \centering
    \includegraphics[width=0.95\columnwidth]{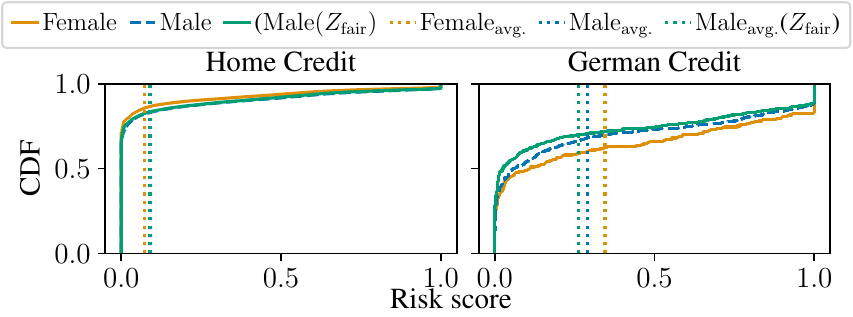}
    \caption{\textbf{Fair risk score estimation in S.2} 
    for females and males under different treatment distributions $\pi(Z)$. Unfair estimates use the factual distribution; fair estimates apply an interventional distribution, treating all applicants as female.}
    \label{fig:mitigation}
\end{figure}

\textbf{Fair risk score estimation.}
Following Sec.~\ref{sec: fair-risk}, we now analyze \emph{treatment-fair} risk score estimates.
Based on insights from scenario S.2, we focus on male applicants and apply different treatment distributions $\pi(Z)$ to generate risk estimates.
We approximate $\pi(Z)$ for males under two conditions: (i) a fair interventional distribution, derived from the empirical distribution of $Z$ conditioned on female applicants to produce ``fair'' risk scores, and (ii) the factual ``unfair'' distribution, based on the empirical distribution of $Z$ conditioned on male applicants.
For female applicants, $\pi(Z)$ is approximated using the empirical distribution of $Z$ conditioned on female applicants. We %
compare the empirical cumulative distribution functions (CDFs) of the resulting risk scores for male and female applicants in Fig.~\ref{fig:mitigation}.

\textbf{Takeaway.} 
While males generally exhibit lower risk scores than females for German Credit, the pattern reverses for Home Credit. 
However, historical treatment practices have led to an \textit{unfair overestimation} of males’ risk scores, especially in German Credit. 
This bias is corrected when the risk scores are recalculated under the fair interventional distribution, adjusting for potentially unfair treatments. 
Integrating these fair risk scores into decision-making pipelines mitigates the residual impacts of historical treatment biases, paving the way for more fair decision-making processes.

\section{Discussion}\label{sec:discussion}

Through a novel causal framework, our work highlighted the overlooked role of non-binary treatment decisions in fairness analyses. This section notes several open challenges (additional discussion in Appendix~\ref{apx:assumptions}).

\textbf{Robust causal frameworks.} 
Our approach assumes access to ground-truth outcomes and no hidden confounders, assumptions that often fail in practice. Hidden confounders~\cite{kilbertus2020sensitivity} and proxy labels for unmeasured outcomes~\cite{mhasawade2024causal} can bias fairness measures, particularly when different stakeholders (e.g., decision-makers vs. auditors) apply the framework. Incorporating emerging methods~\cite{almodovar2025decaflow} is a key step toward addressing these limitations.

\textbf{Considering diverse stakeholders.}
Although our pre-processing approach mitigates treatment discrimination, it lacks theoretical guarantees of optimality. Future work should benchmark our method against existing interventions~\cite{coston2020counterfactual}. Moreover, such approaches may not satisfy differing stakeholder objectives, raising a central question: \textit{What defines a fair treatment policy that accounts for all stakeholders?} For instance, banks may prioritize repayment, while applicants seek lower interest rates~\cite{o2020near}. Achieving fairness thus requires a \textit{holistic} approach to learning policies that balance these utilities.

\textbf{Collecting treatment data.} 
While our framework generalizes to various domains (Appendix~\ref{app:diff_domains}), we evaluated only in lending owing to major data challenges in other domains.
Many datasets conflate covariates with treatments, omit treatment details (e.g., COMPAS~\cite{angwin2022machine} lacks bail terms; lending datasets exclude interest rates), or oversimplify complex treatments into binary variables (e.g., IHDP~\cite{madras2019fairness} reduces care to a single binary indicator). Moreover, most datasets are subject to selective labeling~\cite{lakkaraju2017selective}, where outcomes are observed only when both decision-makers and individuals agree on terms, ignoring potential \emph{agreement-phase discrimination}. These limitations underscore the need for transparent data collection across domains that includes full treatment assignments~\cite{CfPD2023JusticeDataGaps} and agreement-phase decisions to enable more comprehensive analyses.

\textbf{Conclusion.}
Our work underscores the importance of analyzing the fairness of non-binary treatment decisions given positive binary decisions, e.g., loan approvals. Our novel treatment disparity measures extend existing fairness frameworks toward a more comprehensive evaluation of algorithmic decision-making. 
Our findings show that fairness cannot rely on technical solutions alone; it requires \textit{socio-technical} approaches that account for diverse stakeholder utilities.
By capturing the nuanced impacts of treatment decisions, our framework advances aligning algorithmic systems with societal values, working toward \textit{jointly addressing} fairness in binary predictions and non-binary treatment decisions.

\section*{Acknowledgements}
We thank Adrián Javaloy for providing invaluable guidance and feedback regarding the causal generative models. 
This work has been funded by the European Union ({ERC-2021-STG, SAML, 101040177}). 
However, the views and opinions expressed are those of the author(s) only and do not necessarily reflect those of the European Union or the European Research Council Executive Agency. Neither the European Union nor the granting authority can be held responsible.

\bibliography{references}
\clearpage
\appendix

\section{Extended Related Work} \label{apx:relatedwork}

\textbf{Fairness in lending.}
Fairness analyses in automated lending decisions have focused mainly on statistical methods~\cite{hardt2016equality,hurlin2022fairness,liu2018}, emphasizing metrics like equality of opportunity, as noted in surveys~\cite{kozodoi2022fairness,fuster2022predictably}. However, these approaches often overlook biases embedded in the \emph{ground-truth labels}, which stem from unfair treatment during loan processing. While prior work has separately examined treatment discrimination~\cite{sackey2018gender,agier2010credit} and biased outcomes in lending decisions~\cite{munnell1996mortgage, alesina2013women,montoya2020bad}, the causal link between \emph{biased treatment} and \emph{adverse outcomes} remains underexplored. 
Our novel framework bridges this gap and jointly analyzes treatment discrimination and its causal effects on lending outcomes.

\textbf{Causal notions of predictive fairness.}
Causal predictive fairness notions examine the \emph{causal} influence of sensitive features on decisions and outcomes.  
\citet{kusner2017counterfactual} initiated the study of the \emph{total causal effect} of sensitive attributes, followed by work on path-specific effects~\cite{nabi2018fair,chiappa2019path}. Subsequent research extended this to multi-level features~\cite{mhasawade2021causal} and outcome fairness~\cite{plecko2023causal}. However, these approaches largely focused on \emph{binary decisions or outcomes}, overlooking decision-makers' complex \emph{treatment decisions}. While~\citet {madras2019fairness} proposed a causal treatment fairness framework, it reduced non-binary treatments to binary variables, which limits its expressiveness. We address this gap by incorporating the decision-maker’s perspective to provide a fine-grained analysis of treatment decisions and their causal effects. Furthermore, %
we improve the \textit{path-specific counterfactual accuracy} in societal contexts using the \emph{state-of-the-art} CNFs~\cite{javaloy2024causal}, outperforming earlier variational approaches~\cite{chiappa2019path, madras2019fairness}.

\textbf{Multi-stage selection problems.}
Multi-stage selection problems~\cite{emelianov2019price,jia2024learning} explore automating decision-making with binary decisions \textit{across stages}, revealing \textit{additional candidate covariates} for subsequent decisions. Concerning this line of work, our causal framework addresses a fundamentally different context governed by different mechanisms.
Our work focuses on a \emph{single-stage} decision process where binary decisions (e.g., loan approval) are made in one step and \textit{treatment decisions} are directly controlled by decision-makers. We complement existing studies by analyzing the fairness of assigned multivariate, non-binary treatments and their causal impact on decision outcomes.

\section{Methodology Details}

\subsection{Treatment Decisions Across Different Domains}
\label{app:diff_domains}
\begin{table*}[t]
\centering  
\renewcommand{\arraystretch}{1.1}
\resizebox{\textwidth}{!}{
\begin{tabular}{l|lll|l|l}
\toprule
\textbf{Domain} & \textbf{Sensitives $S$} & \textbf{Covariates $X$} & \multicolumn{1}{l}{\textbf{\begin{tabular}[c]{@{}l@{}}Treatment\\ Decisions $Z$\end{tabular}}} & 
\multicolumn{1}{|l|}{{\color[HTML]{848484}{\begin{tabular}[c]{@{}l@{}}Binary\\ Decisions\end{tabular}}}} & \multicolumn{1}{l}{\textbf{Outcomes $Y$}} \\
\midrule
Lending & gender, age, race & \begin{tabular}[c]{@{}l@{}}salary, savings, \\ employment history...\end{tabular} & \begin{tabular}[c]{@{}l@{}}loan terms (amount, \\ duration, interest rate...)\end{tabular} & {\color[HTML]{848484} loan approved} & loan repaid \\ \midrule
Justice & \begin{tabular}[c]{@{}l@{}}gender, race, age, \\ socioeconomic status\end{tabular} & \begin{tabular}[c]{@{}l@{}}criminal history, \\ type of crime...\end{tabular} & \begin{tabular}[c]{@{}l@{}}bail terms (bail amount, \\ conditions...)\end{tabular} & {\color[HTML]{848484} bail approved} & re-offence \\ \midrule
Health & \begin{tabular}[c]{@{}l@{}}gender, age, \\ race, disability\end{tabular} & \begin{tabular}[c]{@{}l@{}}medical history, \\ lab results...\end{tabular} & \begin{tabular}[c]{@{}l@{}}medical treatment \\ (medication, surgery...)\end{tabular} & {\color[HTML]{848484} hospitalization} & re-hospitalization \\ \midrule
Insurance & gender, age, race & \begin{tabular}[c]{@{}l@{}}goods price, salary,\\ employment history...\end{tabular} & \begin{tabular}[c]{@{}l@{}}insurance terms \\ (cost, coverage...)\end{tabular} & {\color[HTML]{848484} {policy approved} } & claim approval \\ \midrule
Hiring & gender, age, race & \begin{tabular}[c]{@{}l@{}}education, experience,\\ qualifications...\end{tabular} & \begin{tabular}[c]{@{}l@{}}employment terms \\ (working hours, salary...)\end{tabular} & {\color[HTML]{848484} job offered} & job performance \\ \bottomrule
\end{tabular}
}
 \caption{
    Examples of data collected across various decision-making domains, highlighting the distinction between sensitive attributes $S$, decision-subject covariates $X$, treatment decisions $Z$, and outcomes $Y$. The multivariate, non-binary $Z$ have been represented previously in the literature as a positive binary decision (in {\color[HTML]{848484}{gray}}).
}
\label{table:identification_of_features}
\end{table*}

Table~\ref{table:identification_of_features} details treatment decisions and covariates across various domains relevant to algorithmic fairness. For each domain, we identify the treatment decision features, which are those that are under the decision-maker's control and directly influence the outcome. The table also specifies the binary decision for each domain. Importantly, these binary decisions may serve as either ground truth labels or proxies for the underlying outcome of interest. For example, the criminal justice domain: while automated systems typically predict bail approval decisions, the true objective is assessing defendant risk. Judges use this risk assessment to make the binary bail decision and consequently determine bail terms. However, focusing solely on the fairness of the binary decisions of bail approval might fail to address the unfairness in the bail terms.

Existing research in the social sciences and medicine has shown potential biases in treatment decisions. For instance, legal scholars have shown potential bias issues in bail treatment decisions~\cite{arnold2018racial}, while some quantitative research has shown unexplained gender biases in insurance costs~\cite{pernagallo2024women}. Similarly, \citet{castilla2008gender} empirically analyzed potential gender disparities in salary hikes in organizations. In healthcare, potential biases were also shown to exist in several treatment factors~\cite{hershman2005racial}.

Our proposed \emph{causal framework} can be readily used to study treatment disparities and outcome effects across \emph{all these representative decision-making domains}.
Importantly, our framework can allow for a more comprehensive evaluation of such real-world scenarios using state-of-the-art computational methods.
However, accessing appropriate datasets across most domains remains a significant challenge, as stated by existing research~\cite{pernagallo2024women}. Even when some data is available, they lack sufficient fine-grained information about treatment decisions provided by the decision-makers.
With future advancements in data availability, our proposed causal framework can be used to study potential treatment disparities and fairness issues across multiple domains.

\subsection{Background on Causal Normalizing Flows} \label{app:CNF}

Causal normalizing flows (CNF)~\cite{javaloy2024causal} enable causal inference from observational data under minor assumptions. 
It provides an ideal practical methodology to estimate unknown causal relations from observational data and an accurate technique to estimate interventional and counterfactual quantities.
Here, we provide a brief background on CNFs and refer the reader to the original paper for full details.

\subsubsection{Assumptions}
The work assumes the following for its theoretical grounding. First, it assumes a \emph{diffeomorphic data generation process}. Hence, the structural equation functions $F$ in the SCM are invertible, and the functions and their inverse are differentiable. Second, it assumes the causal mechanism has \textit{no feedback loops}; hence, it can be represented by a \emph{directed acyclic graph} (DAG). Third, it assumes \textit{causal sufficiency}. Hence, the exogenous variables for each feature are mutually independent, i.e., $p(U) = \prod_ip(u_i)$. Finally, it assumes that a causal ordering is given across the features.

\subsubsection{Autoregressive architecture}
Under the assumptions, the authors show that SCM mechanisms correspond to a special class of autoregressive functions known as \textbf{triangular monotone increasing (TMI) maps}. Through TMI maps, the SCM is represented by triangular, invertible functions. Each feature in $V$ is a monotone function of its parents $pa(V)$ and its exogenous noise term $u$. These maps can be estimated using \emph{autoregressive normalizing flows (ANFs)}, where each conditional is modeled as a monotone transformation, forming the causal normalizing flows model.

The CNF defines a parameterized function $T_\psi$ that maps exogenous variables $U$ to observed features $V$, serving as a generative model for the features. Leveraging the autoregressive structure, the inverse $T_\psi^{-1}$ maps $V$ back to the exogenous space $U$. Since we assume the {causal topological order is known}, each conditional in the flow corresponds directly to a structural equation in the SCM.

By modeling the SCM in this way, CNFs can flexibly approximate the joint distribution over observed variables while remaining consistent with the underlying causal structure. For further details on the architecture and training procedure, we refer to the main paper~\cite{javaloy2024causal}.

\subsubsection{Discussion on Theoretical Guarantees}
Under the stated assumptions and leveraging the flow-based approximation of the SCM, the authors show that if the CNF $(T_\psi, p(U))$ reproduces the observational distribution generated by the true SCM, then the exogenous variables learned by the flow differ from the true exogenous noise variables only by component-wise invertible transformations. Specifically, for each feature $i$, the corresponding exogenous factor learned by the CNF is a transformation $h_i(u_i)$ of the true exogenous variable $u_i$, where $h_i$ is an invertible function. This result establishes that CNFs guarantee \emph{identifiability of exogenous variables up to invertible transformations}, and by successfully isolating the exogenous information, they \emph{ensure causal consistency} in the learned generative process.

\subsubsection{Interventions and \texttt{do}-operation}
While the traditional three-step process of Pearl~\cite{pearl2000models} manipulates the features directly for interventions, CNFs perform interventions by manipulating the latent exogenous variables.
The process is discussed in Section~\ref{sec:implementation}. For generating counterfactuals, we first use the flows model to map the observed features $\vartheta^F$ to exogenous factors as $u = T_\psi(\vartheta^F)$. Next, for the intervention $\doit(V_j \rightarrow \alpha)$, we recompute the exogenous factor \textit{specifically for feature} $j$, hence, $u_j=T_\psi(\doit(V_j \rightarrow \alpha))$. Hence, in the exogenous space,  only for the exact intervened feature, the exogenous variable is updated. For all other features, they remain the same. Finally, one uses the inverse of the flows model $T^{-1}_\psi$ to go from post-interventional $u$ to $\vartheta^{CF}$. Further details on the $\doit$-operation in the CNF can be found in Appendix C in~\cite{javaloy2024causal}.

\subsubsection{Handling partial causal graphs}
A key strength of CNFs, which we leverage in our work, is their ability to learn from \emph{partial causal graphs}, where the complete causal graph is unknown, but the \textit{causal ordering between subsets of observed features} is available. Each subset of features with unknown internal ordering can be treated as a \textit{block}, referred to in~\cite{javaloy2024causal} as a \textit{strongly connected component (SCC)}. 

Given knowledge of the causal ordering between these blocks, the CNF applies autoregressive flow modeling across the blocks. Within each block, the joint distribution is modeled without assuming any specific ordering. Consequently, the CNF retains its theoretical guarantees and identifiability properties \textit{at the block level}, while identifiability is not ensured \textit{within individual blocks}.

In our causal framework, we adopt this formulation by treating all covariates $X$ and all treatment decisions $Z$ as separate blocks. Through our counterfactual analysis of treatment and outcome disparities, we evaluate \textit{block-level counterfactuals} for covariates and treatments to study their causal effects on the downstream outcome $Y$.

\subsection{Extending to Multiple Sensitive Attributes}
\label{app:def_multisens}

Our definitions for measuring treatment disparities and their outcome effects (Definitions~\ref{def:ttd}-\ref{def:dte_y}) can be naturally extended to scenarios where we have \emph{multi-valued} and \emph{multiple} sensitive attributes, e.g, non-binary gender identities, race/ethnicity, religion, etc.

Let us consider $K$ different sensitive attributes, where each attribute $S_i$ has \emph{unique values} $\Scal_i$. Note that this naturally considers binary and multi-valued categorical sensitive attributes.
The combined set of all possible sensitive attribute values can be denoted as $\Scal=\Scal_1 \times \Scal_2 \times \dots \times \Scal_K$.
To compute \emph{counterfactual treatment decisions and outcomes}, a central challenge is interpreting what constitutes a \emph{meaningful counterfactual} across multiple sensitive attributes. Multiple plausible counterfactuals may exist (depending on which attributes are considered), and we outline two different strategies here.

\begin{itemize}
   \item \textbf{Single flip}. We change one sensitive attribute at a time (e.g., only gender), holding the others fixed to their factual values. For example, what treatment would an individual receive if their gender were different but their race and religion stayed the same? For attribute $S_i$, the counterfactual treatment can be defined for any value $s'_i$ as:
    \[\zscf = \zcf(\doit(S_i \rightarrow s'_i)), \forall \, s'_i \in \Scal_i \setminus \{s^F_i\} \]

    \item \textbf{Joint flip}.  We simultaneously change all sensitive attributes to a new disadvantaged group (changing the value set of $S$ from $s^F$ to a different value set $s'$). For example, an individual who is factually a White male Christian would be compared to a Black female Muslim. Formally, the counterfactual treatment would be:
    \[\zscf = \zcf(\doit(S \rightarrow s')), \forall \, s' \in \Scal \setminus \{s^F\} \]
\end{itemize}
Our framework accommodates both counterfactual strategies, enabling interventions $\doit(S \rightarrow s')$ across multiple sensitive attributes using the estimated CNF model. However, each choice corresponds to a distinct fairness question of interest, whether one is concerned with the effects of individual attributes or disparities across entire groups. The work of~\cite{wastvedt2024intersectional, alvarez2025counterfactual} defines counterfactuals involving multiple sensitive attributes through a joint flip, whereas~\cite{yang2020causal} defines counterfactuals relative to a chosen reference subgroup.

\paragraph{Treatment disparities.}
Since we now need to consider potentially multiple sensitive counterfactuals across different sensitive attributes and values, we have to extend our definitions to aggregate disparities properly. For treatment-level disparities (Definitions~\ref{def:ttd} and~\ref{def:dtd}), we can apply an aggregation operator $\Acal$ over the absolute treatment differences $\big|\Delta(\zscf_{s'}, z^F)\big|$ before taking the expectation over the dataset $\Dcal$. For instance, the multi-valued TTD would be:
\begin{align}
\label{eq:ttd_multi}
\operatorname{TTD}_{\text{M}}(s^F) 
= \mathbb{E}_{\Dcal} \Big[ 
\Acal \big( \{ |\Delta(\zscf_{s'}, z^F)| \}_{s' \in \Scal \setminus \{s^F\}} \big) 
\Big].
\end{align}
Following~\cite{wastvedt2024intersectional}, we can consider different types of aggregate functions.
\begin{enumerate}
    \item \textbf{Average}: \[\Acal_{\text{AVG}}(s^F) = \frac{1}{(| \Scal | - 1)| \Scal| /2} \sum_{s' \in \Scal \setminus \{s^F\}} \big|\Delta (\zscf_{s'}, z^F)\big|\]
    \item \textbf{Maximum absolute error}:
    \[\Acal_{\text{MAX}}(s^F) = \max_{s' \in \Scal \setminus \{s^F\} } \big|\Delta (\zscf_{s'}, z^F)\big|\]
    \item \textbf{Variance}:
    \[\Acal_{\text{VAR}}(s^F) =\frac{1}{(| \Scal | - 1)| \Scal| /2} 
    \sum_{s' \in \Scal \setminus \{s^F\} } \Delta(\zscf_{s'}, z^F)^2\]
\end{enumerate}
Definition~\ref{def:dtd} can be extended similarly by computing path-specific counterfactuals $\zpscf$ instead of $\zscf$ for all $\Scal \setminus \{s^F\}$ and using the proposed aggregation methods.

\paragraph{Outcome disparities.}  
We first collect the flips in the outcome label $Y$ caused by the counterfactual treatments:
\[
Y_{\text{flip}} = \Big\{ \mathbb{I}[y^{CF}(\doit(Z \rightarrow \zscf_{s'})) \neq y^F] \;\mid\; s' \in \Scal \setminus \{s^F\} \Big\}.
\]

We then apply an outcome-level aggregation operator $\Acal_Y$ over these indicators to summarize the disparity for the individual, and finally take the expectation over the dataset:
\begin{align}
\label{eq:tte_y_multi}
\operatorname{TTD\text{-}E}_{\text{M}}(s^F, y^F) 
= \mathbb{E}_{\Dcal} \Big[ \Acal_Y(Y_{\text{flip}}) \Big].
\end{align}
For the \emph{worst-case} outcome disparity, $\Acal_Y$ is the identity applied to the indicator corresponding to the treatment with the maximum $\Delta$ (corresponding to $\operatorname{arg\,max}$ over the counterfactual treatment assignments).  
For other aggregations like \emph{average} or \emph{variance} in counterfactual outcomes, $\Acal_Y$ would be the corresponding aggregation function ($\operatorname{mean}$ or $\operatorname{Var}$).
One can use this idea to extend Definition~\ref{def:dte_y} similarly to compute the direct effects across all sensitive values $\Scal$.

\paragraph{Intersectionality.} As discussed, our framework supports analysis involving multiple sensitive attributes. However, sensitive attributes may not always act independently, hence their effects can interact in additive, non-additive, or even compounded ways, influencing treatment and outcome disparities in complex patterns~\cite{crenshaw2013demarginalizing, kanubala2025misalignment, xenidis2018multiple, makkonen2002multiple}. In particular, disparities arising from intersectionality \textit{go beyond} simple multi-sensitive attribute analysis. Intersectionality involves interdependent and interlocking systems of social influence and oppression, where the combined effect of attributes cannot be fully decomposed into separate or additive components~\cite{crenshaw2013demarginalizing}. As such, capturing such intersectional sources of unfairness requires richer causal representations and remains an important direction for future work.

\begin{figure*}[t!]
    \centering
    \begin{subfigure}[t]{0.8\textwidth}
        \centering
        \includegraphics[width=\textwidth]{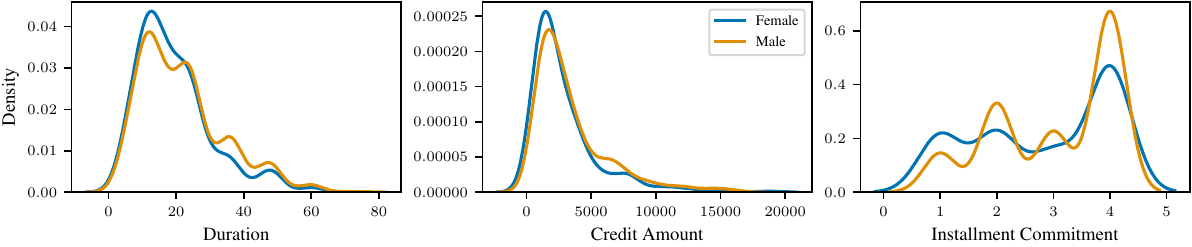} 
        \caption{German Credit %
        }
    \end{subfigure}
    \begin{subfigure}[t]{0.8\textwidth}
        \centering
        \includegraphics[width=\textwidth]{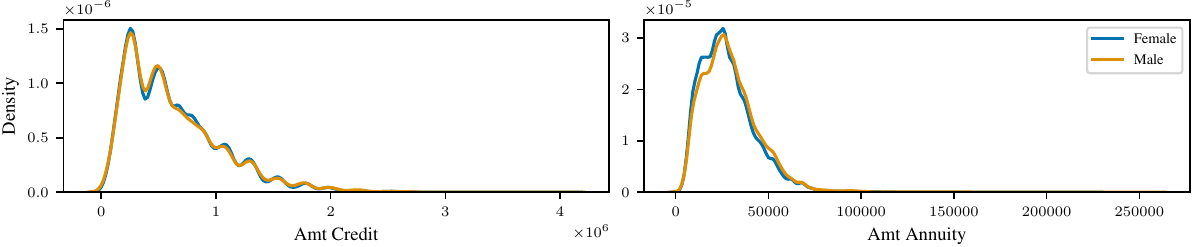}
        \caption{Home Credit}
    \end{subfigure}
    \caption{Treatment decision feature distributions across social groups show overlap, indicating it as a reasonable causal assumption for real-world settings.
    }
    \label{fig:overlap}
\end{figure*}

\subsection{Extended Discussion on Assumptions, Limitations, and Practical Considerations}
\label{apx:assumptions}

We provide an extended discussion on the implicit assumptions of our existing causal framework and practical considerations for deployment.

\paragraph{Partial causal knowledge.}
Our approach requires only \emph{partial knowledge} of the underlying causal mechanisms—an assumption that is \emph{significantly weaker} than that of many existing causal fairness methods~\cite{kusner2017counterfactual,chiappa2019path}, which typically assume full access to the causal graph. Specifically, we assume only the causal ordering among $S$, $X$, $Z$, and $Y$, as depicted in Figure~\ref{fig:reformulated-graph}. Both $X$ and $Z$ are treated as feature blocks, and we \emph{do not require} knowledge of the internal causal ordering among covariates or treatment components. By leveraging CNFs to learn causal relations directly from observed data, our framework preserves theoretical guarantees even under partial causal knowledge, thereby improving its practical applicability.

\paragraph{No hidden confounders.}
Consistent with prior work in causal inference~\cite{khemakhem2021causal,sanchez2022vaca,javaloy2024causal} and causal fairness~\cite{kusner2017counterfactual,chiappa2019path}, we assume a data-generating process with \emph{no hidden confounders}. From the \emph{decision-maker’s perspective}, this corresponds to having complete access to the treatments $Z$, covariates $X$, and sensitive attributes $S$ used to assign $Z$, thereby removing confounding between $(Z,Y)$ and $(X,Z)$.
In practice, however, decision-makers may lack full knowledge of all relevant covariates $X$ influencing $Y$, introducing hidden confounding between $X$ and $Y$. Such confounding \emph{does not affect} our metrics for \textit{total disparities} (Definitions~\ref{def:ttd} and~\ref{def:tte_y}) but \emph{can bias} our \textit{direct disparity measures} (Definitions~\ref{def:dtd} and~\ref{def:dte_y}), which rely on interventions on $X$.

\paragraph{External auditing and confounding.}
Under an \emph{external auditor perspective}, transparency in treatment assignment may be limited, and some components of $Z$ or the covariates $X$ used to determine $Z$ might be unobserved, introducing confounding among $X$, $Z$, and $Y$. Existing studies~\cite{kilbertus2020sensitivity,byun2024auditing} provide fairness bounds under confounding, but they rely on strong assumptions or do not accommodate multivariate, non-binary treatments. Other approaches~\cite{xu2021learning,wang2021proxy,miao2023identifying} employ instrumental or proxy variables to mitigate hidden confounding, yet they are restricted to population-level effects and \emph{lack the individual-level specificity} required for counterfactual analysis. This underscores the need for further research on fairness under confounding and the development of robust, individualized counterfactual methods. Incorporating recent techniques that enable reliable causal inference under \emph{specific types of confounding}~\cite{almodovar2025decaflow} into our framework would be a promising direction for future work.

\paragraph{Overlap.}
Beyond confounding, another key assumption underlying our framework and CNFs is \emph{overlap}~\cite{oberst2020characterization}. We assume that the intervened values in the SCM have \textit{non-zero support} within the original data-generating process. This assumption generally holds in practice. For instance, our considered real-world lending datasets show substantial overlap in treatment distributions across sensitive groups. As seen in Figure~\ref{fig:overlap}, we have significant overlap between the distributions of treatment decisions for female and male applicants. Hence, assigning treatment decisions of the sensitive counterfactual is a realistic consideration for these datasets. Nevertheless, certain real-world contexts may exhibit \emph{lack of overlap} between treatment distributions, for which extending our causal framework remains an important avenue for future research.

\paragraph{Counterfactual validity.}
Accurate fairness evaluation depends on the fidelity of generated counterfactuals. Under valid causal assumptions, CNFs can produce highly accurate counterfactuals~\cite{javaloy2024causal}. However, validating these assumptions in real-world environments remains challenging. Our framework offers a practical advantage in this regard: since counterfactuals in our setting involve \emph{only treatment interventions} — which are under the decision-maker’s control — they can be empirically evaluated via interventional studies. By deploying counterfactual treatment policies to specific subpopulations and measuring the resulting average effects, practitioners can assess and refine the quality of counterfactuals and the causal framework in practice.

\paragraph{Mitigation approaches.}
Our mitigation strategy focuses on \emph{pre-processing} to address treatment unfairness. Such techniques have proven effective in binary predictive fairness~\cite{kamiran2012data} and provide a natural foundation for mitigating treatment disparities. However, pre-processing methods may not suit all scenarios and often lack theoretical guarantees. Future work should explore \emph{in-processing} strategies capable of jointly optimizing fairness and utility objectives. Designing such approaches is non-trivial, as fair policy learning must account for binary decision fairness, non-binary treatment fairness, and stakeholder utility optimization. Achieving these objectives likely requires \emph{multi-objective optimization}~\cite{padh2021addressing} for holistic fairness. Moreover, since treatment decisions and outcomes are typically conditioned on prior binary decisions (e.g., loan approvals), data may exhibit \emph{selection bias}~\cite{lakkaraju2017selective}. Addressing this challenge may require online learning or policy adaptation methods~\cite{kilbertus2020fair,rateike2022don}. Finally, future work should also assess whether in-processing methods can fully meet the multi-dimensional fairness requirements in decision-making~\cite{favier2025cherry}, or whether entirely new paradigms are needed.

\subsection{Defining Stakeholder Losses}
\label{sec:append-losses}
Here, we define the stakeholder loss functions we utilize to measure the utilities of decision-makers and decision-subjects for our mitigation strategy.

\paragraph{Loss Given Default (LGD):}
We estimate the loss incurred by the bank in the event an applicant defaults. Quantifying the loss will require additional analysis of other variables, such as the value of collateral, 
 instalments payments etc~\cite{investopedia2024}. There are several different ways to estimate $LGD$; however, a common approach considers the exposure at default and the recovery rate which is:
\begin{align}
\textbf{LGD}  & = (1 - \text{recovery rate}) \times  \text{exposure at default (EAD)}]
\end{align}
In our estimation, we assume that EAD is equivalent to the credit amount loaned out. EAD is a crucial metric that estimates the total loss a bank is exposed to when a loan defaults. However, it is important to note that the calculation of EAD can be more complex, as it may involve factors beyond the simple loan amount, such as commitment details~\cite{baselIIrisk2012}.

The recovery rate, on the other hand, is a risk-adjusted measure that reflects the portion of the loan that the bank expects to recover after default. In our analysis, we considered it as the outcome of the loan (i.e., the actual repayment).

Finally, we formalize the expected loss given default for the bank as:
\begin{align}
\textbf{LGD}  & = \mathbb{E}_{\mathcal{D}'} [ (1-y ) \times \text{amount} ]
\end{align}
where $y$ represents the applicant's repayment probability.

\paragraph{Expected Simple Interest (ESI):}
We estimate the loss of the individuals using the simple interest formula:

$$\textbf{SI} = \frac{(\text{amount} \times \text{rate} \times \text{time})} {100}$$

For the Home Credit dataset, we used slightly modified formulae to estimate the ESI for Home Credit, as the dataset did not include information on the rate and time, making ESI estimation less straightforward.
We estimate ESI using annuity and loan amount as follows: %
we assume $\text{duration}$=15 years. Annuity can be written as : 
\begin{equation*}
    \text{annuity} = \frac{(\text{amount + interest})}{\text{total durations(months)}} = \frac{(\text{amount + interest})}{12 \times \text{duration}}
\end{equation*}

Using this the simple interest can be calculated as : 
\begin{align}
\textbf{ESI} \text{(Home Credit)} = \mathbb{E}_{\mathcal{D}'}[\text{annuity} \times 12 \times \text{duration} -\text{amount}]
\end{align}

In our formulae for the German Credit dataset, the loan duration is in months and we assume in our implementation that everyone was given the same interest rate of 10\%, hence enabling us to rewrite the ESI as follows:
\begin{align}
\textbf{ESI}\text{(German Credit)} = \mathbb{E}_{\mathcal{D}'} \left[  y \times \text{amount} \times \frac{\text{rate}}{100} \times \frac{\text{time}}{12}    \right]
\end{align}

\subsection{Counterfactual Computation Pseudocodes}
\label{apx:pseudocodes}

This section provides the different function pseudocode for computing the various counterfactual and path-specific counterfactual entities using the causal normalizing flows.
\emph{It is important to note and ensure that whenever we are setting a categorical feature value for intervention, the value is clamped to be within range and floored.}

The algorithm names provided below \textit{match} the function names used in the source code for easy understanding. Moreover, note that for our default consideration of binary sensitive attributes, $s^{CF}$ can be considered as $(1-s^F)$ following the notation in the main paper.

\subsubsection{Generating sensitive counterfactual}
Algorithm~\ref{alg:sens_cf} computes the sensitive counterfactual following the causal relationships when $\doit(S\rightarrow s^{CF})$.
For this computation, we perform the single $\doit$ operation on $S$ as $\doit(S \rightarrow s^{CF})$.
\begin{algorithm}[t]
\caption{\texttt{compute\_sensitive\_counterfactual}}
\label{alg:sens_cf}
\textbf{Input}: Factual data $\Dcal^F$\\
\textbf{Parameters}: Flows \texttt{model}, data  \texttt{preparator}, sensitive index $i_S$ (default: 0) \\
\textbf{Output}: Sensitive counterfactual data $\Dcal^{SCF}(\doit(S \rightarrow s^{CF}))$ 
\begin{algorithmic}[1]
    \State Get sensitive feature values $s^F$ from $\Dcal^F$.
    \State Get sensitive counterfactual feature value $s^{SCF}$ for intervention. 
    \State Get data \texttt{scaler} from \texttt{preparator}. 
    \State $\Dcal^{SCF}$ $\gets$ \texttt{model.compute\_CF} ($\Dcal^F$, $i_S$, $s^{SCF}$, \texttt{scaler})
    \State \textbf{return} $\Dcal^{SCF}$. \Comment{clamp and floor categorical features before saving.} 
\end{algorithmic}
\end{algorithm}

\subsubsection{Computing downstream effect of counterfactual treatments}
Algorithm~\ref{alg:ps_treatment_interv} helps to compute the downstream effect on $Y$ if we perform $\doit(Z \rightarrow \hat{z})$, where $\hat{z}$ is \emph{some interventional value}.
This function is used to (i) understand the downstream effect of Total Treatment Discrimination (TTD) and (ii) understand the downstream effect of Direct Treatment Discrimination (DTD).

\begin{algorithm}[t]
\caption{ \texttt{compute\_path\_specific\_downstream\_}\\\texttt{effect\_treatment\_intervention}}
\label{alg:ps_treatment_interv}
\textbf{Input}: Factual data $\Dcal^F$, counterfactual data $\hat{\Dcal}$ \\
\textbf{Parameters}: Flows \texttt{model}, data  \texttt{preparator} \\
\textbf{Output}: $\Dcal(s^F,x^F,\doit({Z \rightarrow \hat{z}}))$
\begin{algorithmic}[1]
    \State Get treatment indices $i_{\{Z\}}$ from \texttt{preparator} 
    \State Get values $\hat{z}$ from $\hat{\Dcal}$ using $i_{\{Z\}}$ 
    \State Get \texttt{scaler} from \texttt{preparator} 
    \State Clamp and floor $\hat{z}$ categorical features to proper value ranges 
    \State $\Dcal^F(\doit(Z \rightarrow \hat{z}))$ = \texttt{model}.\texttt{compute\_CF}($\Dcal^F$, $i_{\{Z\}}$, $\hat{z}$, \texttt{scaler})
    \State return $\Dcal(S^F,X^F,\doit(Z \rightarrow \hat{z}))$  \Comment{Extract $Y(s^F,x^F,\doit(Z \rightarrow \hat{z}))$ and clamp, floor.}
\end{algorithmic}
\end{algorithm}

\subsubsection{Computing direct effect of sensitive on the target}

Algorithm~\ref{alg:ps_sens_targ} computes the direct effect $S$ on $Y$, i.e. $\doit(S \rightarrow s^{CF})$ on the target $Y$ by computing $y(\doit(S \rightarrow s^{CF}), \doit (X,Z \rightarrow x^F, z^F))$.

\begin{algorithm}[t]
\caption{\texttt{compute\_path\_specific\_sensitive}\\\texttt{\_direct\_label}}
\label{alg:ps_sens_targ}
\textbf{Input}: Factual data $\Dcal^F$, sensitive counterfactual data $\Dcal^{SCF}$ from Algorithm~\ref{alg:sens_cf} \\
\textbf{Parameters}: Flows \texttt{model}, data  \texttt{preparator} \\
\textbf{Output}: $\Dcal(\doit(S \rightarrow s^{CF}), \doit (X,Z \rightarrow x^F, z^F))$
\begin{algorithmic}[1]
    \State Get \texttt{scaler} from \texttt{preparator} 
    \State $U^F$ = \texttt{model.flow.transform}($\Dcal^{SCF}$) \Comment{Helps get $U^F_Y$}
    \State $U^{SCF}$ = \texttt{model.flow.transform} ($\Dcal^{SCF}$) \Comment{Helps get $U^{SCF}_S$}
    \State Get intervention indices $i_{\{X,T\}}$
    \State Clamp, floor categorical dimensions in $\Dcal^F[i_{\{X,T\}}]$, save in \texttt{intervention\_vals} \Comment{Pre-apply \texttt{clone} on $\Dcal^F$}
    \State Set \texttt{data\_tmp} = \texttt{clone}($\Dcal^{SCF}$)
    \State Set \texttt{data\_tmp}$[i_{\{X,T\}}]$ = \texttt{intervention\_vals}$[i_{\{X,T\}}]$
    \State $U^{\mathcal{I}}$ = \texttt{model.flow.transform} (\texttt{data\_tmp}) \Comment{Helps get $U^{\mathcal{I}}_{\{X,Z\}}$}
    \State Get combined $U_{\text{comb}}=[U^{SCF}_S,U^{\mathcal{I}}_{\{X,Z\}},U^F_Y]$
    \State $\Dcal(\doit(S \rightarrow s^{CF}), \doit (X \rightarrow x^F, Z \rightarrow z^F))$ = \texttt{model.flow.transform.inverse}($U_{\text{comb}}$)
    \State return $\Dcal(\doit(S \rightarrow s^{CF}), \doit (X,Z \rightarrow x^F, z^F))$ \Comment{Extract $y(\doit(S \rightarrow s^{CF}), \doit (X,Z \rightarrow x^F, z^F))$ and clamp, floor.}
\end{algorithmic}
\end{algorithm}

\subsubsection{Computing path-specific treatment}
Algorithm~\ref{alg:treatment_ps} calculates the path-specific counterfactuals i.e., the path-specific treatment $z^{PS}=z(\doit (S \rightarrow s^{CF}), \doit (X \rightarrow x^F))$.

\begin{algorithm}[t]
\caption{\texttt{compute\_path\_specific\_treatment}}
\label{alg:treatment_ps}
\textbf{Input}: Factual data $\Dcal^F$, sensitive counterfactual data $\Dcal^{SCF}$ from Algorithm~\ref{alg:sens_cf} \\
\textbf{Parameters}: Flows \texttt{model}, data  \texttt{preparator} \\
\textbf{Output}: $\Dcal^{PS}=D(\doit (S \rightarrow s^{CF}), \doit (X \rightarrow x^F))$
\begin{algorithmic}[1]
\State Get \texttt{scaler} from \texttt{preparator}
\State $U^F$ = \texttt{model.flow.transform}($\Dcal^{SCF}$) \Comment{Helps get $U^F_Y, U^F_Z$}
\State $U^{SCF}$ = \texttt{model.flow.transform} ($\Dcal^{SCF}$) \Comment{Helps get $U^{SCF}_S$}
\State Get intervention indices $i_{X}$
\State Clamp, floor categorical dimensions in $\Dcal^F[i_X]$, save in \texttt{intervention\_vals} \Comment{Pre-apply \texttt{clone} on $\Dcal^F$}
\State Set \texttt{data\_tmp} = \texttt{clone}($\Dcal^{SCF}$)
\State Set \texttt{data\_tmp}[$i_X$] = \texttt{intervention\_vals}[$i_X$]
\State $U^{\mathcal{I}}$ = \texttt{model.flow.transform}(\texttt{data\_tmp}) \Comment{Helps get $U^{\mathcal{I}}_{X}$}
\State Get combined $U_{\text{comb}} = [U^{SCF}_S, U^{\mathcal{I}}_X, U^F_Z, U^F_Y]$
\State $\Dcal^{PS}$ = \texttt{model.flow.transform.inverse}($U_{\text{comb}}$)
\State return $\Dcal^{PS}$ \Comment{Extract $z^{PS}$, clamp and floor categorical features before saving.}
\end{algorithmic}
\end{algorithm}
\paragraph{Computing downstream effect of $z^{PS}$}
Note that once we have $z^{PS}$ from Algorithm~\ref{alg:treatment_ps}, we can compute the downstream effect, i.e., $Y(s^F, x^F, \doit (Z \rightarrow z^{PS}))$ using Algorithm~\ref{alg:ps_treatment_interv}.

\section{Experimental Setup} \label{app:setup}

\subsection{Datasets and Preprocessing}
\label{app:datadetails}

\begin{itemize}
    \item \textit{HMDA:} This dataset regarding mortgage applications covers several states in the United States, but we focused on just two states (New York and Texas) in the year 2017. Following similar preprocessing steps as done in~\cite{cooper2024arbitrariness}, we binarized the gender column as $ \{Male:1, Female:0\}$. Also, we binarized the race with originally $8$ possible values but utilized only $ \{Native White: 0, White: 1\}$. For the action taken with $8$ possible values, we filtered out ${3-8}$. We kept the outcome as a $1$ Loan originated to mean the applicant and the bank decision agreed,  and $2$, which is the application approved but not accepted, to mean a disagreement. In dealing with missing values we removed columns with all NaNs and information regarding the co-applicants. After the preprocessing, we had a total dataset of size $399354$ with $17$ features.

    \item \textit{Home Credit:} We downloaded this dataset from the kaggle competition. We used the application training datasets and selected features for our study based on the feature importance 
    as done in~\cite{matthyspredicting}. We utilized the resulting features selected by their random forest classifier and information entropy models. Based on this, we selected the following features : 
    ``EXT SOURCE 2, EXT SOURCE 3, EXT SOURCE 1, DAYS BIRTH, NAME EDUCATION TYPE, CODE GENDER, DAYS EMPLOYED, NAME INCOME TYPE, ORGANIZATION TYPE, AMT CREDIT, AMT ANNUITY, AMT GOODS PRICE, REGION POPULATION RELATIVE, TARGET".
Based on this subset of data, we dropped the missing values in the dataset and converted the days of birth and days of employment to years using $\frac{\text{days of birth/employment}}{-365}$. We divided it by $-365$ because these values were negative. The complete dataset has a size of $98859$ with $14$ features.

\item \textit{German Credit:} We obtained this dataset from the Statlog website and restructured the personal status feature by consolidating the information on gender and marital status into binary gender categories. Specifically, we grouped male single and male divorced/separated as male, and female divorced/separated and female single as female. The dataset had no missing values. 
\end{itemize}

\textbf{Remarks.} We converted all categorical features using label encoding. In Table~\ref{tab:loan-datasets}, we summarize our datasets by identifying treatment features, sensitive attributes, and outcomes. We note that the treatment decision features for all the lending datasets in our work are \emph{continuous features}. Hence, we use the difference between treatment values as $\Fcal$ and the expectation $\Acal$ over each dataset $\Dcal$ when considering disparity measures in Definitions~\ref{def:ttd} and~\ref{def:dtd}.

\begin{table*}[h!]
\small
\centering
\begin{tabular}{lllll}
\toprule
\textbf{Dataset} &  \begin{tabular}[c]{@{}l@{}}\textbf{Sensitive} \\ \textbf{Attribute (S)}\end{tabular} & \begin{tabular}[c]{@{}l@{}}\textbf{Treatment} \\ \textbf{Decisions (Z)}\end{tabular}  & \textbf{Outcomes (Y)} & \textbf{Dataset Size} \\ \midrule

\begin{tabular}[c]{@{}l@{}}\textbf{HMDA-TX}
\end{tabular}
& \begin{tabular}[c]{@{}l@{}}Gender, Race\\ and Ethnicity\\
\end{tabular} & \begin{tabular}[c]{@{}l@{}}Preapproval,\\ Loan amount\end{tabular} & \begin{tabular}[c]{@{}l@{}}Loan accepted or\\ rejected by the applicant\end{tabular} & ($399354, 17$) \\ \midrule

\begin{tabular}[c]{@{}l@{}}\textbf{HMDA-NY}
\end{tabular} & \begin{tabular}[c]{@{}l@{}}Gender, Race\\ and Ethnicity\\ \end{tabular} & \begin{tabular}[c]{@{}l@{}}Preapproval,\\ Loan amount\end{tabular} & \begin{tabular}[c]{@{}l@{}}Loan accepted or\\ rejected by the applicant\end{tabular} & ($166637, 17$) \\ \midrule

\textbf{Home Credit} & \begin{tabular}[c]{@{}l@{}}Gender,\\ Days of birth (Age)\end{tabular} & \begin{tabular}[c]{@{}l@{}}Annuity,\\ Credit Amount\end{tabular} & Loan repayment & ($98859, 14$) \\ 
\midrule

\textbf{German Credit} & Gender, Age & \begin{tabular}[c]{@{}l@{}}Duration,\\ Credit Amount,\\ Installment rate\end{tabular} & \begin{tabular}[c]{@{}l@{}}Loan repayment\end{tabular} & ($1000, 21$) \\ 
\bottomrule
\end{tabular}
\caption{Different loan approval datasets we used in the study with their sensitive attributes ($S$), treatment decisions ($Z$), outcome label ($Y$) and dataset size.} %
\label{tab:loan-datasets}
\end{table*}

\subsection{Hyper-Parameter Tuning and Training of Causal Normalising Flows}

For the causal normalizing flows~\cite{javaloy2024causal},
we use the original work as inspiration and tune for the flow models' inner dimensionality and learning rate Table~\ref{tab:hyperparam}. All other parameters are
fixed, which are as follows: Adam optimizer with
plateau scheduler, activation as ELU, layer type was Neural Spline Flow (NSF) and all the experiments were run for a maximum of $1000$ epochs. For other parameters, we refer to the GitHub code \footnote{https://github.com/psanch21/causal-flows}. The number of nodes is specified in Table~\ref{tab:feat_cnf}. An example of causal graph input to causal normalizing flows is given in Figure~\ref{fig: train_graph}. 

\begin{table}[ht!]
\small
    \centering
    \begin{tabular}{c|c|c|c}
    \toprule
       \textbf{Dataset}  & \textbf{dim-inner} & \textbf{lr} & \textbf{log-prob}   \\
       \midrule
        \textbf{HMDA-NY} & $32\times32$ & $10^{-2}$ &-61.67 \\
        \midrule
        \textbf{HMDA-TX} & $32\times32$ & $10^{-2}$ &-63.49 \\
        \midrule
        \textbf{Home Credit} & $64$ & $10^{-2}$ &-41.87\\
        \midrule
        \textbf{German Credit} & $64$ & $10^{-3}$ &-14.68 \\
        \midrule
        \textbf{Synthetic Loan} & $16\times16\times16\times16$ & $10^{-3}$ &-16.51\\
    \bottomrule
    \end{tabular}
     \caption{Hyperparameters for causal Flows. %
    The log-prob
        of the observational data distribution estimate was used to
select the best hyperparameters.}
    \label{tab:hyperparam}
\end{table}

\begin{table}[ht!]
    \centering
    \small
    \begin{tabular}{c|c|c|c}
    \toprule
       \textbf{Dataset}  & \textbf{\#rootnodes} & \textbf{\#covariates} & \textbf{\#treatment}  \\
       \midrule
     \textbf{HMDA-NY}    & 2 & 12 & 2\\
         \midrule
         \textbf{HMDA-TX}   & 2 & 12 & 2\\
           \midrule
         \textbf{Home Credit}    & 2  & 9  & 2 \\
             \midrule
            \textbf{German Credit}    & 2  & 15 & 3\\
               \midrule
            \textbf{Synthetic Loan}    & 2 & 3 & 2 \\
        \bottomrule 
    \end{tabular}
      \caption{Number of root nodes, covariates, and treatment variables for training causal normalizing flows model. The dimensionality of the outcome label is 1. }
    \label{tab:feat_cnf}
\end{table}

\subsection{Hardware}

All experiments of hyper-parameter tuning and training of causal normalizing flows are done on an Ubuntu workstation with an Intel CPU. The code primarily uses PyTorch CPU and Python. We set the seed for PyTorch and Python to a fixed number and ran all experiments. A fixed seed is also used to split datasets into train, validation, and test sets - 80\%, 10\%, 10\%. This seed is fed as input in a parameter file to train and run evaluations.

\begin{figure*}[t!]
        \centering
        \resizebox{0.65\linewidth}{!}{
        \begin{tikzpicture}[node distance=2.5cm]

        \node (A) [startstop] {\textbf{A}};
        \node (G) [startstop, below of=A] {\textbf{G}}; %
        \node (hammer) [left=-0.2cm of G,yshift =0.5cm] {\scalebox{2.2}{\textcolor{teal}{\faIcon{hammer}}}};
        
        \node (E) [process, right of=A, xshift=2cm, yshift=1cm] {\textbf{E}};
        \node (I) [process, below of=E] {\textbf{I}};
        \node (S) [process, below of=I] {\textbf{S}};

        \node (L) [process, right of=E, xshift=2cm, yshift=-1cm] {\textbf{L}};
        \node (D) [process, below of=L] {\textbf{D}};

        \node (Y) [startstop, right of=L, xshift=1.5cm, yshift=-1cm] {\textbf{Y}};

        \draw[black, dashed, rounded corners] ($(A.north west)+(-0.75,0.75)$) rectangle ($(G.south east)+(0.75,-0.75)$);
        
        \draw[black, thick, rounded corners] ($(E.north west)+(-0.75,0.75)$) rectangle ($(S.south east)+(0.75,-0.75)$);
        \draw[black, thick, rounded corners] ($(L.north west)+(-0.75,0.75)$) rectangle ($(D.south east)+(0.75,-0.75)$);

        \draw [arrow] (A.east) -- ++(0.75,0) -- ++(0,0) |- ($(E.west) - (0.5, 1)$); %
        \draw [arrow] (G.east) -- ++(0.75,0) -- ++(0,0) |- ($(I.west) - (0.5, 1)$); %

        \draw [arrow] ($(I.east) + (0.5, 0.0)$) -- ++(0.5,0) -- ++(0.5,0) |- ($(L.west) - (0.45,1.5)$);%
        \draw [arrow] ($(L.east) + (0.5, -1)$) -- ++(0.5,0) -- ++(0.75,0) |- (Y.west); %

        \draw [arrow] ($(A.north west)+(2,-0.10)$) ++(-1.0,0.8) to[out=65,in=90] (Y.north);
        \draw [arrow] ($(G.south east)+(1.0,-2.7)$)  ++(-2.0,2.0) to[out=-60,in=-125] ($(D.south west)+(0.5,-0.75)$);

        \draw [dashedarrow] (E) -- (I);
        \draw [dashedarrow] (I) -- (S);
        \draw [dashedarrow] (L) -- (D);

        \node[align=center] at ($(G.south) + (0,-3.2)$) {\textbf{Sensitive $S$}};
        \node[align=center] at ($(Y.south) + (0.2,-4.7)$) {\textbf{Outcome $Y$}};
        \node[align=center] at ($(S.south) + (0,-1.7)$) {\textbf{Block $X$}};
        \node[align=center] at ($(D.south) + (0,-3.2)$) {\textbf{Block $Z$}};

        \end{tikzpicture}
        }
        \caption{Causal graph for synthetic loan dataset introduced in Appendix~\ref{app:synthetic}. While training the causal normalising flows, covariates $X$ and treatment decisions $Z$ are treated as multivariate blocks without any explicit assumptions on the cause-effect directionality within each block. Here, we have multiple sensitive attributes (independent root nodes) but for this work, we concentrate on intervening on just one sensitive attribute (``gender''). A similar analysis can be done by intervening on the attribute ``age''.}
        \label{fig: train_graph}
    \end{figure*}

\section{Synthetic Validation Evaluation} \label{app:synthetic}
In this section, we consider a synthetic data generation process where we have access to the ground-truth generative process. Using this dataset, we aim to validate the performance of our causal framework, comparing it to the \emph{ground-truth oracle} that has access to the exact causal relations. This analysis shows that, considering the underlying assumptions hold, our causal framework leveraging the CNF can capture unknown causal relations and reliably help audit existing treatment disparities and also mitigate potential discriminatory effects of such treatment decision disparities.
\subsection{Synthetic Dataset Generation}
Taking inspiration from \cite{karimi2020algorithmic,majumdar2024carma}, we consider the 7-variable loan dataset
modeling the credit dataset \cite{hofmann1994statlog} with an added outcome variable $Y$ (Equation~\ref{eq:synth}). Similar to \cite{karimi2020algorithmic,majumdar2024carma}, the equations are non-linear with non-additive exogenous variables. Gender $G$ and age $A$ are root-nodes, loan
amount $L$, education-level $E$, income $I$, and savings $S$ are actionable with equations shown below. Additionally, we added a link between the savings $S$ to the loan amount and thresholded the outcome variable $Y$ directly on $G$ using two different thresholding scenarios. $\sigma(\cdot)$ represents the
sigmoid function in the following equations. $\Ncal (a,b)$
denotes the Gaussian distribution where a is the mean and b is the variance of the distribution, $\sqrt{b}$ denoting the standard deviation. Finally, $\operatorname{Gamma}(a, b)$ denotes the Gamma distribution where a is the concentration/shape and b is the scale, 1/b being the rate.

\begin{figure}[t]
\centering
\fbox{%
\begin{adjustbox}{max width=\columnwidth}
\small
\begin{minipage}{1.0\columnwidth}
\[
\begin{aligned}
&\textbf{Root nodes :} \\
    &f_G : G =U_{G}, \quad   U_{G}~\sim \operatorname{Bernoulli}(0.5)\\
    &f_A : A=-35+U_{A},\quad    U_{A}~\sim \operatorname{Gamma}(10,3.5)\\
 &\textbf{Covariates:} \\   
    &f_E : E=-0.5+\sigma(-1+0.5 G+\sigma(0.1 A)+U_{E}),\\ & \quad \quad U_{E}\sim\mathcal{N}(0,0.25)\\
    &f_I : I=-4+0.1(A+35)+2 G+G E+U_{I},
     U_{I}~\sim~\mathcal{N}(0,4)\\
    &f_S : S=-4+ 1.5 \Ibb\{I > 0\} I+U_{S},
    U_{S}~\sim~\mathcal{N}(0,5) \\
& \textbf{Treatments:} \\ 
    &f_L : L= 1+0.01(A-5)(5-A)+ 2(1-G) + \beta S  +U_{L}, \\
    & \quad \quad U_{L}~\sim~\mathcal{N}(0,10), \beta \in \{0, 0.03\} \\
    &f_D : D=-1+0.1 A + 3 (1-G) +L+U_{D}, U_{D}~\sim~\mathcal{N}(0,9)\\
& \textbf{Outcome: two possibilities} \\     
     &f_Y : Y=\Ibb\{\sigma(\delta (-L -D) + 0.3 (I+ S + \alpha I  S ) ) \\
     &\geq 0.5  + \gamma (1-G)\} \quad  \textbf{OR}  \\
      &f_Y : Y=\Ibb\{\sigma(\delta (-L -D) + 0.3 (I+ S + \alpha I  S ) \\
      & + U_Y* \gamma (1-G)) \geq 0.5 \} , U_{Y}~\sim~\mathcal{N}(0,5)\\
&\textbf{ With direct effect S to Y if } \gamma>0 , \textbf{where} \\ 
& \alpha = 1 , \textbf{if} \left[\Ibb\{I>0\} \land \Ibb\{S>0\})\right]=1, \\ & \alpha=-1, \textbf{otherwise} \\ 
& \textbf{where} \quad \Ibb \quad \textbf{is an indicator function.}\\
\end{aligned}
\]
\end{minipage}
\end{adjustbox}
}
\caption{Synthetic data generating process.}
\label{eq:synth}
\end{figure}

\begin{table*}[ht!]
    \centering
    \resizebox{\textwidth}{!}{
    \renewcommand{\arraystretch}{1.0}
    \begin{tabular}{l c c |c c c c |c c c c}
        \toprule
        \multirow{2}{*}{\textbf{Dataset}} & \multirow{2}{*}{\textbf{Sensitive counterfactual}} & \multirow{2}{*}{\textbf{Measure}} & \multicolumn{4}{c|}{\textbf{Treatment Disparity}} & \multicolumn{4}{c}{\textbf{Effect on Repayment (\%)}} \\
        
        \cmidrule(lr){4-7} \cmidrule(lr){8-11}
        & & & \multicolumn{2}{c}{\textbf{Flows}} & \multicolumn{2}{c|}{\textbf{Oracle}} & \multicolumn{2}{c}{\textbf{Flows}} & \multicolumn{2}{c}{\textbf{Oracle}} \\
         \cmidrule(lr){4-7} \cmidrule(lr){8-11}
         & & & \textbf{Amount} & \textbf{Duration} & \textbf{Amount} & \textbf{Duration} &\boldmath{$Y^F = 0$} & \boldmath{$Y^F = 1$} & \boldmath{$Y^F = 0$} & \boldmath{$Y^F = 1$} \\
        
        \midrule
        \multirow{5}{*}{\textbf{Balanced}} & \multirow{2}{*}{Female $\rightarrow$ Male} & TTD(-E) & $-1.91$ & $-4.44$ & $-1.95$ & $-4.94$& $0.19$ & $0.00$ & $0.18$ & $0.00$ \\
    & & DTD(-E) & $-1.62$ & $-4.25$ & $-2.00$& $-5.00$ & $0.17$ & $0.00$& $0.18$ & $0.00$ \\
   \cmidrule(lr){2-11}
    & \multirow{2}{*}{Male $\rightarrow$ Female} & TTD(-E) & $+2.14$ &  $+4.64$ &$+1.95$ &$+4.95$& $0.00$ & $0.20$ & $0.00$ & $0.20$\\
    & & DTD(-E) & $+1.90$ &   $+4.05$ &$+2.00$ &$+5.00$ & $0.00$ & $0.17$ & $0.00$ & $0.20$\\
   \midrule
    \multirow{4}{*}{\textbf{Imbalanced}} & \multirow{2}{*}{Female $\rightarrow$ Male} & TTD(-E) & $-1.92$ &  $-4.61$ & $-1.95$ & $-4.95$ & $0.09$& $0.00$ & $0.09$ & $0.00$\\
    & & DTD(-E) & $-1.55$ &  $-3.75$ & $-2.00$ &$-5.00$ & $0.07$ & $0.00$ & $0.09$ & $0.0$ \\
   \cmidrule(lr){2-11}
    & \multirow{2}{*}{Male $\rightarrow$ Female} & TTD(-E) & $+2.00$ &  $+4.69$ & $+1.95$ & $+4.95$ & $0.00$ & $0.21$ & $0.00$ & $0.21$\\
    & & DTD(-E) & $+2.21$ & $+5.45$ & $+2.00$ & $+5.00$ & $0.00$& $0.24$ & $0.00$ & $0.21$ \\
    \bottomrule
 \end{tabular}}
\caption{\textbf{Measuring treatment disparities and discrimination for synthetic data.} Oracle uses perfect causal knowledge. 
Here, we have loan amount and duration as the treatment features. Results are reported as median values. \textbf{Our framework using CNF can accurately capture disparities without access to ground-truth causal mechanisms.}}
\label{tab:syntheticdatasets}
 \end{table*}

 \begin{table}[ht!]
    \small
    \begin{tabular}{lcccc}
    \toprule 
    \textbf{Sens. Group} & \multicolumn{2}{c}{\textbf{Balanced Data}} & \multicolumn{2}{c}{\textbf{Unbalanced Data}} \\ 
    \cmidrule(lr){2-3} \cmidrule(lr){4-5}
    & $\textbf{LGD}$  & $\textbf{ESI}$  & $\textbf{LGD}$  & $\textbf{ESI}$  \\  
    \midrule
    Males & $3.09$ & $0.58$ & $3.30$ & $0.58$ \\
    Females ($\Dcal$) & $4.14$ & $0.36$ & $3.49$ & $0.30$ \\
    Females ($\Dcal^\mathrm{fair}$) & \textbf{3.14} & $0.55$ & \textbf{2.39} & $0.47$ \\
    \bottomrule 
    \end{tabular}
     \centering
    \caption{\textbf{Fairness in non-binary decision-making.} Comparison of the bank's LGD and the applicant's ESI losses (in DM) for balanced and unbalanced datasets under different treatments.}
    \label{tab:combinedlosses} 
\end{table}
 
\subsection{Results}
Using the synthetic data generation process emulating the lending scenario as shown above, we generate balanced and unbalanced datasets and repeat our empirical analysis to i) use our novel notions to analyze 
the data for treatment discrimination, and ii) explore our mitigation strategies to enable treatment-fair decisions. 
Additionally, with access to the ground-truth generative process, we can consider an oracle model with the ground-truth knowledge.
Using the oracle, we can perform interventions to compute the oracle treatment discrimination values. This allows us to compare the results given by our CNF-based causal framework with the oracle.

\subsubsection{Balanced dataset:} 

Using the synthetic loan generation process described in the previous section, we set the parameters $\beta=0.03, \gamma=0.5, \delta=1, \eta=5$ to construct a balanced dataset containing $5000$ samples. The dataset was balanced concerning both gender and the outcome variable $Y$. We evaluated the performance of the causal normalizing flows (CNF) model by comparing its results to the oracle model that leverages perfect causal knowledge. The comparison, detailed in Table~\ref{tab:syntheticdatasets}, demonstrates that the CNF model effectively captures the underlying structure of the observational data. This corroborates that CNF is a robust and reliable model for learning from observational data while accounting for causal relationships.

\textit{Results:}
From Table~\ref{tab:syntheticdatasets}, we observe that females receive longer loan durations and larger loan amounts compared to their male counterparts. 
This significant discrimination is reflected in both $\operatorname{TTD}$ and $\operatorname{DTD}$. Remarkably, treating females as males would lead to a $0.19\%$ of defaulting females to repay, while no female applicant would have defaulted under this treatment. This suggests that a \emph{positive treatment discrimination} exists towards females.

To address this discrimination, it might be beneficial to apply the treatment of males to their female counterparts. This adjustment aims to mitigate treatment discrimination and improve repayment outcomes. Using this insight, we generate the fair dataset $\mathcal{D^\text{fair}}$ as shown in Section~\ref{sec: fair-nonbinary}. We further evaluate this treatment intervention by calculating the bank's loss ($\operatorname{LGD}$) and applicant loss ($\operatorname{ESI}$) with results summarized in Table~\ref{tab:combinedlosses}. Interestingly, while applying male treatment to females reduces the bank's losses, it increases the applicants' losses, highlighting that such an intervention benefits one stakeholder (the bank) but not the female applicants.

Figure~\ref{fig:balanced} illustrates the empirical cumulative distribution function (CDF) of \emph{risk scores} under different treatment distributions 
$\pi(Z)$: the factual distribution and the interventional treatment distribution conditioned on male applicants. This analysis reveals that males typically exhibit lower risk scores than females, with a substantial gap between the two. Applying the suggested fairer treatment reduces this gap, making the risk scores fairer across genders by mitigating the treatment discrimination.

\textit{Takeaway:} 
While our risk score estimation strategy shows improvement in mitigating treatment discrimination, our fairness in the decision-making strategy falls short of improving conditions for all stakeholders in the decision-making setup by merely adjusting the treatment decisions. This limitation likely arises because our current approach lacks guarantees on the optimality of the designed utility functions. 
To address this shortcoming, we need to develop more balanced and fair treatment policies $\pi_z$ that not only mitigate discrimination but also optimize outcomes for all stakeholders involved. Designing such policies will enable a more fair and efficient decision-making pipeline.

\subsubsection{Unbalanced dataset}

To create the imbalanced dataset, we adjusted the parameters to $\beta=0.03$, $\gamma=0.5$, $\delta=2$, and $\eta=5$, generating a dataset with $5000$ samples. This configuration resulted in a dataset with a higher proportion of males compared to females and a greater number of individuals who defaulted on their loans than those who repaid them. These reflect real-world disparities often encountered in financial datasets, providing a more challenging scenario for evaluating the model's performance.

\begin{figure*}[t!]
    \centering
    \begin{subfigure}[t]{0.45\textwidth}
        \centering
        \includegraphics[width=\textwidth]{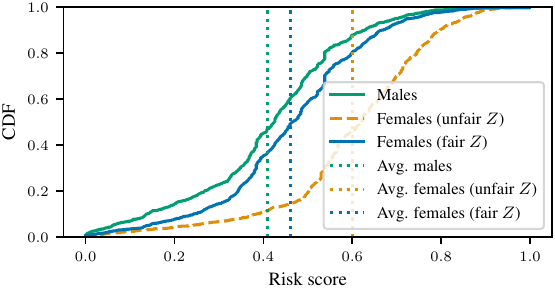} 
        \caption{Balanced %
        }
        \label{fig:balanced}
    \end{subfigure}
    \begin{subfigure}[t]{0.45\textwidth}
        \centering
        \includegraphics[width=\textwidth]{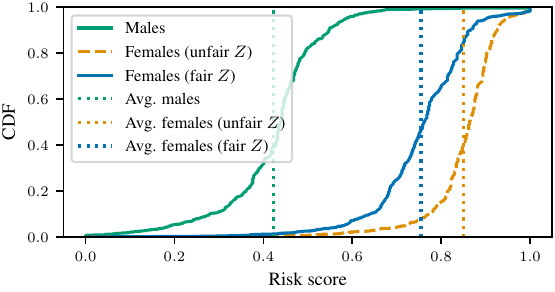}
        \caption{Unbalanced}
        \label{fig:unbalanced}
    \end{subfigure}
    \caption{\textbf{Fair risk score estimation.} 
    Comparison of risk score estimates for females and males under various treatment distributions $\pi(Z)$. Unfair estimates are based on the factual distribution, while fair estimates use interventional treatment distribution where all applicants are treated as males.
    }
    \label{fig:mitigation_synthetic}
\end{figure*}

\textit{Results:} Referring to Table~\ref{tab:syntheticdatasets}, we observe a similar trend to the balanced dataset, females receive longer loan durations and higher credit amounts compared to males. Likewise, there is significant discrimination in the impact of both $\operatorname{TTD}$ and $\operatorname{DTD}$ on repayment, just like in the balanced case. When male treatment decisions are applied to females, the risk scores of females improve, as shown in Figure~\ref{fig:unbalanced} and the gap in the risk score decreases. Additionally, Table~\ref{tab:combinedlosses} highlights a consistent pattern: treating females as males reduces bank losses but increases applicant losses, following the same trend observed in the balanced dataset scenario.

\textit{Takeaway:} 
The unbalanced dataset reveals a similar trend, highlighting the limitations of our current treatment policies $\pi_z$, which may be sub-optimal in addressing fairness and stakeholder utilities. The results highlight the necessity of developing treatment policies that are not only fair but also tailored to optimize utility across diverse groups in the dataset. For the unbalanced dataset, fairness treatment interventions must account for the unequal distribution of data points across groups since the imbalance can make discrimination worse and impact the stakeholder utilities. Designing fair and optimal policies $\pi_z$ for such scenarios requires integrating strategies that address both the imbalance in representation and the need for fair treatment decisions. %

\section{Additional Real-World Evaluations}
\label{app:additional_eval}

\begin{table*}[t!]
 \centering
  \small
 \begin{tabular}{l c c c c c}
    \toprule
     \multirow{2.5}{*}{\textbf{Dataset}} & \multirow{2.5}{*}{{\textbf{\begin{tabular}[c]{@{}l@{}}Sensitive\\ Counterfactual\end{tabular}}}}  & \multirow{2.5}{*}{\textbf{Measure}} & \textbf{Treatment Disparity} & \multicolumn{2}{c}{\textbf{Effect on Agreement($\%$)}}\\
    \cmidrule(lr){4-4} \cmidrule(lr){5-6}
    & & & \textbf{Amount ($ \times 10^3$\$) } &  \boldmath{$Y^F= 0$ }& \boldmath{$  Y^F= 1$ } \\
    \midrule
    \multirow{4}{*}{\textbf{HMDA-NY}} & \multirow{2}{*}{Female $\rightarrow$ Male} & TTD(-E) & $+29.46$ &  $2.00$ & $0.40$ \\
    & & DTD(-E) & $+6.65$ & $2.00$ & $0.20$\\
    \cmidrule(lr){4-4} \cmidrule(lr){5-6}
    & \multirow{2}{*}{Male $\rightarrow$ Female} & TTD(-E) & $-29.93$ & $4.00$ & $0.16$\\
    & & DTD(-E) & $-5.88$ &  $2.00$  & $0.14$\\
    \midrule
    \midrule
    \multirow{4}{*}{\textbf{HMDA-TX}} & \multirow{2}{*}{Female $\rightarrow$ Male} & TTD(-E) & $+19.43$ & $2.00$ & $0.28$\\
    & & DTD(-E) & $+2.67$ & $2.00$ & $0.13$\\
  \cmidrule(lr){4-4} \cmidrule(lr){5-6}
    & \multirow{2}{*}{Male $\rightarrow$ Female} & TTD(-E) & $-19.78$ & $3.00$ & $0.18$ \\
    & & DTD(-E) & $-3.65$ & $3.00$  & $0.10$\\
    \bottomrule
 \end{tabular}
  \caption{
 \textbf{Measuring treatment disparities in S.1 and outcome effects.} {Analysis conducted on the entire, unsplit dataset.} $Y^F =0$ indicates a label change from negative to positive agreement, $Y^F =1$ the reverse. Results show median U.S. Dollars ($\$$). \textbf{Results on entire data consistent with trends in main paper.}
 }
\label{tab:HMDA-apx}
 \end{table*}

\begin{table*}[t!]
\centering
\resizebox{0.87\textwidth}{!}{%
\begin{tabular}{l c c c c c c}
    \toprule
     \multirow{2.5}{*}{\textbf{Dataset}} & \multirow{2.5}{*}{{\textbf{\begin{tabular}[c]{@{}l@{}}Sensitive\\ Counterfactual\end{tabular}}}}  & \multirow{2.5}{*}{\textbf{Measure}} & \multicolumn{2}{c}{\textbf{Treatment Disparity}} & \multicolumn{2}{c}{\textbf{Effect on Repayment($\%$)}}\\
   \cmidrule(lr){4-5}  \cmidrule(lr){6-7}
     & & & \textbf{Annuity (INR)} & \textbf{Amount (INR)} & 
     \boldmath{$ Y^F= 0$} & \boldmath{$ Y^F= 1$} \\
      \midrule
     \multirow{4}{*}{\textbf{Home Credit}} & \multirow{2}{*}{Female $\rightarrow$ Male} & TTD(-E) & $+1008.12$   & $+7885.72$ & $0.24$ & $0.30$ \\
    & & DTD(-E) & $+715.91$ &   $-816.94$  & $0.27$  & $0.00$ \\
    \cmidrule(lr){4-5}  \cmidrule(lr){6-7}
    & \multirow{2}{*}{Male $\rightarrow$ Female} & TTD(-E) & $-942.92$ & $-6659.68$  & $2.30$ & $0.00$\\
    & & DTD(-E) &  $-594.56$  & $+758.46$ & $1.30$  & $0.00$ \\
     \midrule
     \midrule
      & & & \textbf{Duration (months)} & \textbf{Amount (DM)} & \boldmath{$ Y^F=0$} & \boldmath{$Y^F=1$} \\
     \cmidrule(lr){4-7}  %
    \multirow{4}{*}{\textbf{German Credit}} & \multirow{2}{*}{Female $\rightarrow$ Male} & TTD(-E) & $+1.29$ & $+253.34$  & $2.00$ & $3.50$\\
    & & DTD(-E) & $+0.46$ & $+164.66$  & $2.00$ & $3.00$\\
    \cmidrule(lr){4-5}  \cmidrule(lr){6-7}
    & \multirow{2}{*}{Male $\rightarrow$ Female} & TTD(-E) & $-0.79$ & $-243.11$ & $8.00$ & $1.00$ \\
    & & DTD(-E) & $-0.59$ & $-212.39$ & $8.00$ &  $0.60$\\
    \bottomrule
\end{tabular}
}
\caption{
\textbf{Measuring treatment disparities in S.2 and outcome effects}. {Analysis conducted on the entire, unsplit dataset.} $Y^F = 0$ indicates a label change from loan defaulted to repaid, $Y^F = 1$ the reverse. Results show median INR (Indian Rupees) and DM (Deutsche Marks). \textbf{Results on entire data consistent with trends in main paper.}
}
\label{tab:loanrepayment-analysis-apx}
\end{table*}

\subsection{Analyzing Treatment Disparities on the Entire Data}
\label{apx:whole-data-analysis}
We repeat our evaluations conducted in the main paper regarding the auditing of historical treatment disparities and their unfair effect on outcomes. Here, we consider only the \emph{historical decision policy} that gave loans and analyze the whole proportion of each lending dataset.

\textbf{Results in HMDA.} 
We report the results in Table~\ref{tab:HMDA-apx}. We analyze the treatment disparity measures ($\operatorname{TTD}$ and $\operatorname{DTD}$) for factual males and females across the two states.
Our analyses show that female applicants, if treated like their male counterfactuals, would receive a substantial increase in loan amounts, e.g., in NY, approximately $\$$29K more based on $\operatorname{TTD}$ (and $\$$6K based on $\operatorname{DTD}$) on average. 
Conversely, if male applicants were treated like females, their loan amounts would decrease by a similar amount. 
Examining the effects on the outcome \textit{agreement rates}, we observe similar $\operatorname{TTD-E}$ and $\operatorname{DTD-E}$ values for both factual females and males, with agreement rates increasing slightly more when males are treated as females. 
The HMDA-TX dataset shows similar, though less pronounced, treatment disparities compared to NY.

\textbf{Results in Home Credit.} 
We report the results in Table~\ref{tab:loanrepayment-analysis-apx}.
Our analyses show that if females were treated like their male counterfactuals, they would have received higher annuities and loan amounts. 
The reverse pattern would apply to male applicants when treated as females. 
Interestingly, we see \textit{contrasting} trends between \textit{total and direct} disparities regarding loan amounts. 
Total disparity indicates conservative lending practices where female applicants are granted lower loan amounts and annuities. 
In contrast, direct disparity, attributed solely to gender, shows that female applicants receive \textit{lower loan amounts} but \textit{higher annuities}, suggesting elevated interest rates. 
These findings suggest that while banks may have implemented some \textit{affirmative} treatment policies, disparities mediated through covariates influence lending decisions significantly.
Importantly, we see minor downstream effects on outcomes across groups. Hence, the outcome differences \textit{do not justify} the observed treatment disparities against females, particularly regarding $\operatorname{DTD}$, highlighting \emph{potential discrimination}.

\textbf{Results in German Credit.}
From the results in Table~\ref{tab:loanrepayment-analysis-apx}, we observe a similar trend with the bank adopting more conservative practices toward female applicants. 
If females were treated like males, they would receive longer loan durations and higher loan amounts, with the reverse pattern observed for males. 
Significant disparity is seen in loan amounts, driven mainly by direct effects. 
This observation suggests that applicant covariates do not fully explain treatment disparities, and gender may have a direct influence, exposing a potential problematic mechanism. 
Interestingly, there is no disparity in installment rate treatments regarding $\operatorname{TTD}$ or $\operatorname{DTD}$, so we exclude this feature from Table~\ref{tab:loanrepayment-analysis} for clarity.
When examining the impact of treatment disparities on outcomes, a key observation arises: treating males like their female counterparts would result in 8.0\% of defaulting males repaying their loans. 
In contrast, only 1.0\% of male repayers would default under this counterfactual treatment.
Hence, our counterfactual analysis shows that the treatment disparities are not justified by outcomes, indicating \emph{potential discrimination}.

\begin{table}[ht]
\centering
\small
\begin{tabular}{llllll}
\toprule
\textbf{Dataset} & \textbf{Predictor} & \textbf{Accuracy} & \textbf{DP} & \textbf{EOD} \\
\midrule
\multirow{3}{*}{German Credit} & Accuracy & 70 & 0.06 & 0.18  \\
 & DP & 67.5 & 0.03 & 0.17  \\
 & EOD & 66.5 & 0.02 & 0.15  \\
 \midrule
\multirow{3}{*}{Home Credit} & Accuracy & 69.2 & 0.14 & 0.13  \\
 & DP & 70.8 & 0.0 & 0.05  \\
 & EOD & 68.2 & 0.01 & 0.02  \\
 \midrule
\multirow{3}{*}{HMDA-NY} & Accuracy & 74.0 & 0.01 & 0.01  \\
 & DP & 72.0 & 0.00 & 0.00  \\
 & EOD & 72.0 & 0.00 & 0.01  \\
 \midrule
\multirow{3}{*}{HMDA-TX} & Accuracy & 77.4 & 0.01 & 0.01  \\
 & DP & 77.1 & 0.00 & 0.00  \\
 & EOD & 77.1 & 0.00 & 0.00  \\
 \bottomrule
\end{tabular}%
\caption{Accuracy and fairness (demographic parity: DP, equalized odds: EOD) measures on the test-set of unmitigated (Accuracy) and fairness post-processed (DP, EOD) predictors for binary predictions of loan granting.}
\label{tab:pred-acc}
\end{table}

\begin{table*}[t!]
 \centering
\resizebox{0.85\textwidth}{!}{
 \begin{tabular}{l l c c c c c}
    \toprule
     \multirow{2.5}{*}{\textbf{Dataset}} & \multirow{2.5}{*}{\textbf{{\textbf{\begin{tabular}[c]{@{}l@{}}Binary Policy\\(accepted people)\end{tabular}}}}} & \multirow{2.5}{*}{{\textbf{\begin{tabular}[c]{@{}l@{}}Sensitive\\ Counterfactual\end{tabular}}}}  & \multirow{2.5}{*}{\textbf{Measure}} & \textbf{Treatment Disparity} & \multicolumn{2}{c}{\textbf{Effect on Agreement($\%$)}}\\
    \cmidrule(lr){5-5} \cmidrule(lr){6-7}
    & & & & \textbf{Amount ($ \times 10^3$\$) } &  \boldmath{$Y^F= 0$ }& \boldmath{$  Y^F= 1$ } \\
    \midrule
    \multirow{16}{*}{\textbf{HMDA-NY}} & \multirow{4}{*}{{\textbf{\begin{tabular}[c]{@{}l@{}}Historical\\(66655)\end{tabular}}}} & \multirow{2}{*}{Female $\rightarrow$ Male} & TTD(-E) & $+29.68$ &  $2.18$ & $0.41$ \\
    & & & DTD(-E) & $+6.77$ & $2.51$ & $0.20$\\
    \cmidrule(lr){5-5} \cmidrule(lr){6-7}
    & & \multirow{2}{*}{Male $\rightarrow$ Female} & TTD(-E) & $-29.98$ & $3.88$ & $0.15$\\
    & & & DTD(-E) & $-5.87$ &  $1.77$  & $0.14$\\
    \cmidrule{2-7}
    & \multirow{4}{*}{{\textbf{\begin{tabular}[c]{@{}l@{}}Accuracy\\(46027)\end{tabular}}}} & \multirow{2}{*}{Female $\rightarrow$ Male} & TTD(-E) & $+27.92$ &  $4.39$ & $0.09$ \\
    & & & DTD(-E) & $+6.37$ & $5.26$ & $0.06$\\
    \cmidrule(lr){5-5} \cmidrule(lr){6-7}
    & & \multirow{2}{*}{Male $\rightarrow$ Female} & TTD(-E) & $-27.75$ & $2.34$ & $0.05$\\
    & & & DTD(-E) & $-6.50$ &  $2.34$  & $0.03$\\
    \cmidrule{2-7}
    & \multirow{4}{*}{{\textbf{\begin{tabular}[c]{@{}l@{}}Dem. Parity\\(44492)\end{tabular}}}} & \multirow{2}{*}{Female $\rightarrow$ Male} & TTD(-E) & $+28.03$ &  $5.48$ & $0.08$ \\
    & & & DTD(-E) & $+6.52$ & $4.11$ & $0.06$\\
    \cmidrule(lr){5-5} \cmidrule(lr){6-7}
    & & \multirow{2}{*}{Male $\rightarrow$ Female} & TTD(-E) & $-27.68$ & $2.56$ & $0.04$\\
    & & & DTD(-E) & $-6.58$ &  $2.56$  & $0.03$\\
    \cmidrule{2-7}
    & \multirow{4}{*}{{\textbf{\begin{tabular}[c]{@{}l@{}}Eq. Odds\\(44526)\end{tabular}}}} & \multirow{2}{*}{Female $\rightarrow$ Male} & TTD(-E) & $+27.96$ &  $4.82$ & $0.07$ \\
    & & & DTD(-E) & $+6.48$ & $3.61$ & $0.05$\\
    \cmidrule(lr){5-5} \cmidrule(lr){6-7}
    & & \multirow{2}{*}{Male $\rightarrow$ Female} & TTD(-E) & $-27.68$ & $2.56$ & $0.04$\\
    & & & DTD(-E) & $-6.59$ &  $2.56$  & $0.03$\\
    \midrule
    \midrule
    \multirow{16}{*}{\textbf{HMDA-TX}} & \multirow{4}{*}{{\textbf{\begin{tabular}[c]{@{}l@{}}Historical\\(159742)\end{tabular}}}} & \multirow{2}{*}{Female $\rightarrow$ Male} & TTD(-E) & $+19.38$ &  $1.60$ & $0.29$ \\
    & & & DTD(-E) & $+2.70$ & $1.64$ & $0.14$\\
    \cmidrule(lr){5-5} \cmidrule(lr){6-7}
    & & \multirow{2}{*}{Male $\rightarrow$ Female} & TTD(-E) & $-19.81$ & $3.01$ & $0.17$\\
    & & & DTD(-E) & $-3.65$ &  $2.38$  & $0.10$\\
    \cmidrule{2-7}
    & \multirow{4}{*}{{\textbf{\begin{tabular}[c]{@{}l@{}}Accuracy\\(114723)\end{tabular}}}} & \multirow{2}{*}{Female $\rightarrow$ Male} & TTD(-E) & $+20.98$ &  $0.00$ & $0.02$ \\
    & & & DTD(-E) & $+2.95$ & $0.00$ & $0.02$\\
    \cmidrule(lr){5-5} \cmidrule(lr){6-7}
    & & \multirow{2}{*}{Male $\rightarrow$ Female} & TTD(-E) & $-21.31$ & $0.00$ & $0.01$\\
    & & & DTD(-E) & $-3.35$ &  $0.00$  & $0.01$\\
    \cmidrule{2-7}
    & \multirow{4}{*}{{\textbf{\begin{tabular}[c]{@{}l@{}}Dem. Parity\\(114211)\end{tabular}}}} & \multirow{2}{*}{Female $\rightarrow$ Male} & TTD(-E) & $+20.98$ &  $0.00$ & $0.02$ \\
    & & & DTD(-E) & $+2.95$ & $0.00$ & $0.02$\\
    \cmidrule(lr){5-5} \cmidrule(lr){6-7}
    & & \multirow{2}{*}{Male $\rightarrow$ Female} & TTD(-E) & $-21.32$ & $0.00$ & $0.01$\\
    & & & DTD(-E) & $-3.35$ &  $0.00$  & $0.01$\\
    \cmidrule{2-7}
    & \multirow{4}{*}{{\textbf{\begin{tabular}[c]{@{}l@{}}Eq. Odds\\(114155)\end{tabular}}}} & \multirow{2}{*}{Female $\rightarrow$ Male} & TTD(-E) & $+20.98$ &  $0.00$ & $0.02$ \\
    & & & DTD(-E) & $+2.95$ & $0.00$ & $0.02$\\
    \cmidrule(lr){5-5} \cmidrule(lr){6-7}
    & & \multirow{2}{*}{Male $\rightarrow$ Female} & TTD(-E) & $-21.32$ & $0.00$ & $0.01$\\
    & & & DTD(-E) & $-3.35$ &  $0.00$  & $0.01$\\
    \bottomrule
 \end{tabular}
 }
 \caption{
 \textbf{Measuring treatment disparities in S.1 and outcome effects.} $Y^F =0$ indicates a label change from negative to positive agreement, $Y^F =1$ the reverse. Results show median U.S. Dollars ($\$$). \textit{The policy "Historical" uses the whole test set since the data collected under the original policy is for people who were offered a loan. "Accuracy" indicates that an accuracy-optimized predictor is used to give out loans to a subset of the test population. DP and EOD indicate fairness post-processed predictors used for the loan binary decision-making.} \textbf{Ensuring fairness of binary predictions cannot directly mitigate treatment decision disparities.}
 }
\label{tab:HMDA-diffpreds}
 \end{table*}

\begin{table*}[t!]
\centering
\resizebox{0.98\textwidth}{!}{%
\begin{tabular}{l l c c c c c c}
    \toprule
     \multirow{2.5}{*}{\textbf{Dataset}} & \multirow{2.5}{*}{\textbf{{\textbf{\begin{tabular}[c]{@{}l@{}}Binary Policy\\(accepted people)\end{tabular}}}}} &\multirow{2.5}{*}{{\textbf{\begin{tabular}[c]{@{}l@{}}Sensitive\\ Counterfactual\end{tabular}}}}  & \multirow{2.5}{*}{\textbf{Measure}} & \multicolumn{2}{c}{\textbf{Treatment Disparity}} & \multicolumn{2}{c}{\textbf{Effect on Repayment($\%$)}}\\
   \cmidrule(lr){5-6}  \cmidrule(lr){7-8} 
     & & & & \textbf{Annuity (INR)} & \textbf{Amount (INR)} & 
     \boldmath{$ Y^F= 0$} & \boldmath{$ Y^F= 1$}  \\
      \midrule
     \multirow{16}{*}{\textbf{Home Credit}} & \multirow{4}{*}{{\textbf{\begin{tabular}[c]{@{}l@{}}Historical\\(39543)\end{tabular}}}} &\multirow{2}{*}{Female $\rightarrow$ Male} & TTD(-E) & $+1005.26$   & $+7834.47$ & $0.22$ & $0.25$ \\
    & & & DTD(-E) & $+708.24$ &   $-797.40$  & $0.22$  & $0.08$ \\
    \cmidrule(lr){5-6}  \cmidrule(lr){7-8}
    & & \multirow{2}{*}{Male $\rightarrow$ Female} & TTD(-E) & $-957.37$ & $-6778.88$  & $2.46$ & $0.05$\\
    & & & DTD(-E) &  $-604.94$  & $+857.12$ & $1.36$  & $0.08$ \\
    \cmidrule{2-8}
    & \multirow{4}{*}{{\textbf{\begin{tabular}[c]{@{}l@{}}Accuracy\\(26258)\end{tabular}}}} &\multirow{2}{*}{Female $\rightarrow$ Male} & TTD(-E) & $+1024.06$   & $+8527.55$ & $0.34$ & $0.14$ \\
    & & & DTD(-E) & $+684.38$ &   $-946.94$  & $0.00$  & $0.05$ \\
    \cmidrule(lr){5-6}  \cmidrule(lr){7-8}
    & & \multirow{2}{*}{Male $\rightarrow$ Female} & TTD(-E) & $-930.77$ & $-7427.25$  & $3.28$ & $0.03$\\
    & & & DTD(-E) &  $-516.16$  & $+1115.90$ & $1.64$  & $0.03$ \\
    \cmidrule{2-8}
    & \multirow{4}{*}{{\textbf{\begin{tabular}[c]{@{}l@{}}Dem. Parity\\(27072)\end{tabular}}}} &\multirow{2}{*}{Female $\rightarrow$ Male} & TTD(-E) & $+1024.06$   & $+8527.55$ & $0.34$ & $0.14$ \\
    & & & DTD(-E) & $+687.54$ &   $-952.23$  & $0.00$  & $0.05$ \\
    \cmidrule(lr){5-6}  \cmidrule(lr){7-8}
    & & \multirow{2}{*}{Male $\rightarrow$ Female} & TTD(-E) & $-922.00$ & $-7263.01$  & $2.93$ & $0.03$\\
    & & & DTD(-E) &  $-537.34$  & $+1046.03$ & $1.58$  & $0.06$ \\
     \cmidrule{2-8}
      & \multirow{4}{*}{{\textbf{\begin{tabular}[c]{@{}l@{}}Eq. Odds\\(25864)\end{tabular}}}} &\multirow{2}{*}{Female $\rightarrow$ Male} & TTD(-E) & $+1031.86$   & $+8606.90$ & $0.18$ & $0.13$ \\
    & & & DTD(-E) & $+694.17$ &   $-972.88$  & $0.00$  & $0.05$ \\
    \cmidrule(lr){5-6}  \cmidrule(lr){7-8}
    & & \multirow{2}{*}{Male $\rightarrow$ Female} & TTD(-E) & $-924.41$ & $-7310.10$  & $3.08$ & $0.04$\\
    & & & DTD(-E) &  $-529.54$  & $+1069.00$ & $1.54$  & $0.06$ \\
    \midrule
     \midrule
      & & & & \textbf{Duration (months)} & \textbf{Amount (DM)} & \boldmath{$ Y^F=0$} & \boldmath{$Y^F=1$} \\
     \cmidrule(lr){5-8}  %
    \multirow{16}{*}{\textbf{German Credit}} & \multirow{4}{*}{{\textbf{\begin{tabular}[c]{@{}l@{}}Historical\\(400)\end{tabular}}}} & \multirow{2}{*}{Female $\rightarrow$ Male} & TTD(-E) & $+1.15$ & $+272.15$  & $4.25$ & $2.7$\\
    & & & DTD(-E) & $+0.42$ & $+133.39$  & $4.25$ & $1.35$\\
    \cmidrule(lr){5-6}  \cmidrule(lr){7-8}
    & & \multirow{2}{*}{Male $\rightarrow$ Female} & TTD(-E) & $-0.87$ & $-237.97$ & $12.33$ & $0.97$ \\
    & & & DTD(-E) & $-0.58$ & $-189.68$ & $12.33$ &  $0.48$\\
    \cmidrule{2-8}
    & \multirow{4}{*}{{\textbf{\begin{tabular}[c]{@{}l@{}}Accuracy\\(238)\end{tabular}}}} & \multirow{2}{*}{Female $\rightarrow$ Male} & TTD(-E) & $+0.87$ & $+175.69$  & $0.0$ & $3.51$\\
    & & & DTD(-E) & $+0.36$ & $+112.62$  & $0.0$ & $1.75$\\
    \cmidrule(lr){5-6}  \cmidrule(lr){7-8}
    & & \multirow{2}{*}{Male $\rightarrow$ Female} & TTD(-E) & $-0.67$ & $-171.70$ & $6.90$ & $0.70$ \\
    & & & DTD(-E) & $-0.42$ & $-167.69$ & $3.45$ &  $0.70$\\
    \cmidrule{2-8}
    & \multirow{4}{*}{{\textbf{\begin{tabular}[c]{@{}l@{}}Dem. Parity\\(213)\end{tabular}}}} & \multirow{2}{*}{Female $\rightarrow$ Male} & TTD(-E) & $+0.87$ & $+175.69$  & $0.0$ & $3.51$\\
    & & & DTD(-E) & $+0.36$ & $+112.62$  & $0.0$ & $1.75$\\
    \cmidrule(lr){5-6}  \cmidrule(lr){7-8}
    & & \multirow{2}{*}{Male $\rightarrow$ Female} & TTD(-E) & $-0.66$ & $-207.17$ & $8.70$ & $0.81$ \\
    & & & DTD(-E) & $-0.39$ & $-174.30$ & $4.35$ &  $0.81$\\
    \cmidrule{2-8}
    & \multirow{4}{*}{{\textbf{\begin{tabular}[c]{@{}l@{}}Eq. Odds\\(210)\end{tabular}}}} & \multirow{2}{*}{Female $\rightarrow$ Male} & TTD(-E) & $+0.94$ & $+189.89$  & $0.0$ & $3.64$\\
    & & & DTD(-E) & $+0.36$ & $+116.74$  & $0.0$ & $1.82$\\
    \cmidrule(lr){5-6}  \cmidrule(lr){7-8}
    & & \multirow{2}{*}{Male $\rightarrow$ Female} & TTD(-E) & $-0.67$ & $-212.44$ & $9.1$ & $0.81$ \\
    & & & DTD(-E) & $-0.42$ & $-189.68$ & $4.54$ &  $0.81$\\
    \bottomrule
\end{tabular}
}
\caption{
\textbf{Measuring treatment disparities in S.2 and outcome effects}. $Y^F = 0$ indicates a label change from loan defaulted to repaid, $Y^F = 1$ the reverse. Results show median INR (Indian Rupees) and DM (Deutsche Marks). \textit{The policy "Historical" uses the whole test set since the data collected under the original policy is for people who were given the loan. Accuracy indicates that an accuracy-optimized predictor is used to give out loans to a subset of the test population. DP and EOD indicate fairness post-processed predictors used for the loan binary decision-making.} \textbf{Ensuring fairness of binary predictions cannot directly mitigate disparities in treatment decisions.}
}
\label{tab:loanrepayment-analysis-diffpreds}
\end{table*}

\subsection{Complementarity of Treatment and Predictive Fairness Across Various Predictors}
\label{apx:mult-preds}
In this section, we analyze how different binary prediction models used for making binary predictions (giving loans) impact the treatment disparity values, assuming the treatment decision policy remains fixed to the past policy. Through this evaluation, we show that if the treatment decision policy remains the same, simply intervening for the fairness of binary decisions \emph{does not} mitigate disparities that exist in treatment decisions. Here, other than considering the past historical policy that was used to collect the dataset (the entire dataset is for positive lending decisions; we do not have access to data of individuals rejected for a loan), we consider that the binary predictor is applied to simulate making automated binary decisions \emph{lending decisions}~\cite{corbett2017algorithmic}.

\paragraph{Practical considerations.} We split each dataset 60-40 for train-test, then use the training data subset to train a logistic regression predictor that aims to make binary predictions that improve accuracy based on historically observed outcomes $Y$. Then, we use the \texttt{fairlearn}~\cite{weerts2023fairlearn} package to \emph{post-process} this predictor for (i) demographic parity and (ii) equalized odds, given the historical dataset. For post-processing, we set a particular seed to ensure reproducibility when the fairness algorithm uses randomization. For data pre-processing, we one-hot encode categorical features and use standard scaling on continuous features. The train-test data split was performed using stratified splitting regarding the label $Y$. The base logistic regression and latter fairness post-processing were optimized for \emph{balanced accuracy}. The post-processing algorithm followed~\cite{hardt2016equality}, using a grid size of 5000. The different predictive metrics of the trained predictors are shown in Table~\ref{tab:pred-acc}.

\paragraph{S.1 HMDA Analysis}
We provide the results in Table~\ref{tab:HMDA-diffpreds}.
Across the HMDA-NY and HMDA-TX datasets, the Accuracy, Demographic Parity (DP), and Equalized Odds (EOD) policy models show persistent gender-based treatment disparities that remain close in magnitude to those under the Historical policy. In both states, females receive significantly lower loan amounts than males—about \$28K in NY and \$21K in TX under TTD even after applying fairness-aware adjustments. However, the outcome-level effects of these disparities diverge between the two states. In NY, counterfactual analysis reveals that a meaningful percentage of males would have agreed with loan terms had they received female-like treatment up to 2.34\% under the Accuracy model (TTD and DTD), with very low disagreement rates suggesting that female-assigned terms may result in more acceptable or calibrated offers. This agreement gain persists across DP and EOD models, albeit with slight variations, indicating that these fairness adjustments do little to alter the underlying outcome asymmetry. In contrast, the TX dataset exhibits minimal sensitivity in counterfactual agreement under all models beyond Historical. Agreement shifts are essentially zero under Accuracy, DP, and EOD in both directions, suggesting that while treatment disparities still exist, they do not meaningfully affect borrower agreement in Texas. Taken together, these results highlight that fairness-aware models do not eliminate gender-based disparities in treatment, and particularly in NY, the terms typically offered to females may lead to better borrower agreement outcomes—raising important considerations for fairness beyond simple parity in allocation.

\paragraph{S.2 Home Credit Analysis}
We show the results in Table~\ref{tab:loanrepayment-analysis-diffpreds}.
For the Home Credit dataset, all binary policy models show consistent disparities in treatment between male and female applicants. Across models, male applicants consistently receive higher annuities and loan amounts compared to females (e.g., under the Historical model, Female $\rightarrow$ Male counterfactuals show increases of INR1005.26 in annuity and INR7834.47 in amount). However, we also consistently observe that a non-trivial percentage of male applicants would have repaid their loans if given the terms offered to their female counterfactuals. For example, under the Accuracy model, 3.28\% of males would have repaid under female terms ($Y^F=0$), compared to just 0.03\% who would have defaulted ($Y^F=1$), indicating that female-like terms may be more conservative, supporting higher repayment likelihoods. On the other hand, very small percentages of females would have repaid if given male terms (often <0.4\%), indicating that the male treatment regime is not more advantageous in terms of outcomes. It may increase default risk. Overall, the fairness-aware policies (DP and EOD) reduce neither the disparities in loan amounts nor the trends of repayment outcomes. This specific analysis clearly shows that post-hoc binary prediction fairness adjustments do not fully correct for treatment disparities or their unfair impact on outcomes, critically highlighting the complementary nature of binary decision fairness and treatment decision fairness.

\paragraph{S.2 German Credit Analysis}
The results in Table~\ref{tab:loanrepayment-analysis-diffpreds} for the German Credit dataset reflect the trends observed in the main paper: there are systematic differences in treatment terms and their outcome effects across gender. Across the different predictors tested, males are generally offered longer and larger loans. However, regarding repayment, we see that a large proportion of males would have repaid their loans under female-like terms. While binary predictive fairness-aware models (DP, EOD) reduce this repayment gap somewhat (e.g., $Y^F=0$ for males drops to 9.1\% for EOD), the general trend of disparity and its potentially unfair effect on outcomes remains. Moreover, disparity values in treatment terms, e.g., in amount, increase for the EOD-fair predictor compared to a predictor optimized only for accuracy. These results suggest that fairness in lending should not be judged only by parity in binary decisions, but also regarding the treatment terms provided and how well the terms align with successful repayment.

\begin{table*}[t]
\centering
\resizebox{\textwidth}{!}{%
\begin{tabular}{llllllllll|ll}
\toprule
\textbf{Data} & \textbf{Predictor} & \textbf{Train on} & \multicolumn{3}{c}{\textbf{Bin. Pred.}} & \multicolumn{6}{c}{\textbf{Utility}} \\
\cmidrule(lr){7-12}
($Y=1$) &  & \textbf{Pre-proc.} &  &  &  & \multicolumn{2}{c}{\textbf{Female}} & \multicolumn{2}{c}{\textbf{Male}} & \multicolumn{2}{|c}{\textbf{Overall}} \\
\cmidrule(lr){4-6} \cmidrule(lr){7-8} \cmidrule(lr){9-10} \cmidrule(lr){11-12}
 &  & \textbf{(loans)} & \textbf{Accuracy} & \textbf{DP} & \textbf{EOD} & \textbf{LGD} & \textbf{ESI} & \textbf{LGD} & \textbf{ESI} & \textbf{LGD} & \textbf{ESI} \\
 \midrule
\multirow{6}{*}{\begin{tabular}[c]{@{}l@{}}Home\\Credit\\ \xmark: 92.4\%\\ \cmark: 92.5\%\end{tabular}} & \multirow{2}{*}{Accuracy} & \xmark\ (26258) & 69.2 & 0.14 & 0.13 & 631167.83 & 446661.90 & 651661.23 & 4771788.94 & 636986.18 & 4538969.85\\
 &  & \cmark\ (26240) & 69.1 & 0.13 & 0.12 & 630534.25 & 4444490.55 & 641463.72 & 4513145.93 & \textbf{633674.39} & \textbf{4464215.89} \\
 \cmidrule{2-12}
 & \multirow{2}{*}{DP} & \xmark\ (27072) & 70.8 & 0.0 & 0.05 & 633404.55 & 4451011.27 & 632518.60 & 4728754.74 & 633112.83 & 4542463.90 \\
 &  & \cmark\ (27027) & 70.6 & 0.0 & 0.04 & 631624.03 & 4445472.78 & 622713.60 & 4479214.72 & \textbf{628695.43} & \textbf{4456562.79} \\
 \cmidrule{2-12}
 & \multirow{2}{*}{EOD} & \xmark\ (25864) & 68.2 & 0.01 & 0.02 & 635313.03 & 4455243.59 & 639333.13 & 4742956.96 & 636622.24 & 4548941.77 \\
 &  & \cmark\ (25500) & 67.3 & 0.01 & 0.02 & 634544.57 & 4450423.15 & 630188.55 & 4491738.00 & \textbf{633127.24} & \textbf{4463865.90} \\
 \midrule
\multirow{6}{*}{\begin{tabular}[c]{@{}l@{}}German\\Credit\\ \xmark: 70\%\\ \cmark: 71.7\%\end{tabular}} & \multirow{2}{*}{Accuracy} & \xmark\ (238) & 70.0 & 0.06 & 0.18 & 290.79 & 274.62 & 733.49 & 471.94 & 608.86 & 416.39 \\
 &  & \cmark\ (240) & 69.2 & 0.04 & 0.21 & 282.36 & 277.70 & 464.74 & 373.13 & \textbf{412.31} & \textbf{345.70} \\
 \cmidrule{2-12}
 & \multirow{2}{*}{DP} & \xmark\ (213) & 67.5 & 0.03 & 0.17 & 290.79 & 274.62 & 714.79 & 438.54 & 581.42 & 386.98 \\
 &  & \cmark\ (239) & 68.5 & 0.0 & 0.19 & 445.30 & 281.67 & 475.87 & 364.84 & \textbf{466.66} & \textbf{339.78} \\
 \cmidrule{2-12}
 & \multirow{2}{*}{EOD} & \xmark\ (210) & 66.5 & 0.02 & 0.15 & 299.74 & 277.88 & 643.84 & 457.89 & 537.33 & 402.17 \\
 &  & \cmark\ (229) & 67.0 & 0.03 & 0.2 & 290.79 & 271.89 & 483.76 & 355.67 & \textbf{427.30} & \textbf{331.16}\\
 \bottomrule
\end{tabular}%
}
\caption{Analyzing test-set binary prediction metrics and treatment decision downstream utilities of bank and applicants (\textit{mean across female, male, and overall}), comparing training+analysis on original data (\xmark) against our proposed pre-processed treatment-fair-augmented data (\cmark). $Y=1$ proportions and the number of granted loans across datasets remain similar regardless of pre-processing, allowing for comparison. \textbf{Training on pre-processed data keeps similar binary predictive metrics but greatly improves the downstream overall utility of the bank (LGD) and applicants (ESI).}}
\label{tab:train-preproc}
\end{table*}

{
\subsection{Training Binary Predictors on Treatment-Fair Pre-Processed Data}
\label{apx:train-fair}
We evaluate the effects of training on our proposed treatment-fair pre-processed data on both the fairness and utility of downstream binary predictors designed to make binary lending decisions. Specifically, we compare training binary predictors $\hat{Y}=h_y(S,X,Z)$ on the original treatment-unfair data $\Dcal$ and our pre-processed treatment-fair data $\Dcal^{\mathrm{fair}}$. For each data, we split it into a $60-40$ train-test subsets, train (and post-process for fairness), and analyze on the test set. We again consider a logistic regression and use balanced accuracy. For fairness, we perform post-processing for demographic parity (DP) or equalized odds (EOD). The practical considerations mirror those shown in Appendix~\ref{apx:mult-preds}.
}

{
Table~\ref{tab:train-preproc} reports results across the Home Credit and German Credit datasets, highlighting metrics related to predictive performance (accuracy, demographic parity (DP), and equalized odds (EOD)), as well as utility outcomes measured in terms of expected loss to the bank (LGD) and expected interest paid by applicants (ESI) for female applicants, male applicants, and over all applicants. As discussed in the main paper (Sec.~\ref{sec:exp_mitigation-data}), we pre-process the data by augmenting fairer treatment assignments and their outcome effects on $Y$ for the male subgroup since our analyses revealed historically riskier loan terms offered to this group.}
{
\paragraph{Home Credit.} 
Training on pre-processed data does not significantly affect binary predictive performance, supporting its use as a foundation for downstream model training. When the binary predictor is trained solely for accuracy, binary fairness metrics show slight improvement after pre-processing (DP: 0.14 to 0.13, EOD: 0.13 to 0.12), with a negligible drop in accuracy (69.2\% to 69.1\%). Similar trends hold for the DP- and EOD-trained predictors. Crucially, \textit{pre-processing successfully achieves fairer treatments and consistent improvements in downstream utility} across predictors. Specifically, looking at the \emph{overall expected} LGD and ESI across \emph{all applicants}, we see that training the predictor on our pre-processed data \emph{reduces the expected bank loss and the expected interest paid by applicants}, providing benefits for both stakeholders. Looking at each group, we see a negligible impact on female applicants and a major reduction for male applicants (expectedly since we intervene on this group for pre-processed $\Dcal^{\mathrm{fair}}$), resulting in narrower utility gaps between the groups.
}
{
\paragraph{German Credit.}
We observe similar trends where training binary predictors on pre-processed data generally does not significantly affect predictive performance. For example, the Accuracy-trained predictor shows an improvement in demographic parity (DP: 0.06 to 0.04) but a slight worsening in equalized odds (EOD: 0.18 to 0.21), alongside a minor drop in accuracy (70.0\% to 69.2\%). From a utility standpoint, we see our desired effect of mitigating treatment disparities: pre-processing consistently reduces both the bank’s expected losses (LGD) and applicants’ effective interest rates (ESI) overall, reflecting greater efficiency for the lender and fairer loan terms for borrowers. We see similar trends across the different predictors. Moreover, the gaps in LGD and ESI between female and male applicants have substantially reduced. These results indicate the benefit of training on the pre-processed data: the data-driven decision-making system can now provide loans with better terms that benefit all stakeholders.
}

{
\paragraph{Takeaway.}
Pre-processing to correct treatment disparities does not significantly impact binary predictive performance. Still, it substantially improves treatment fairness and mitigates its downstream effects, making it a practical foundation for model training. Our approach specifically targets disparities in treatment and their resulting impact on financial outcomes. Training on pre-processed data consistently reduces both the bank’s expected losses (LGD) and applicants’ effective interest rates (ESI) overall, thereby benefiting all stakeholders. These improvements also lead to narrower utility gaps between female and male applicants, indicating a more equitable distribution of financial burden. While our results show that treatment-fair pre-processed data can be effectively used to train binary predictors with fairness constraints, improvements in treatment disparity do not always translate into gains in predictive fairness metrics, further indicating the \textit{complementary nature of fairness in binary decisions and non-binary treatment decisions}. Future work should explore more nuanced in-processing methods that jointly address both treatment and predictive fairness to enable more comprehensive fairness improvements while accounting for the utilities of diverse stakeholders.
}

\end{document}